\documentclass[11pt]{article}
\usepackage[english]{babel}
\usepackage[utf8x]{inputenc}
\usepackage{amsmath}
\usepackage{amssymb}
\usepackage{graphicx}
\usepackage{etoolbox}
\usepackage{changepage}
\usepackage{titlesec}
\usepackage[parfill]{parskip}
\usepackage[margin=1in]{geometry}
\usepackage{float}
\usepackage{tikz}
\usepackage{nicematrix}
\usepackage{booktabs}
\usepackage{placeins}
\usepackage{wrapfig}
\usepackage{microtype}
\usepackage{url}
\usepackage{color, colortbl}

\newcommand{\norm}[1]{\left\lVert #1 \right\rVert}

\titleformat*{\section}{\large\bfseries}
\titleformat*{\subsection}{\normalsize\bfseries}

\usepackage{listings,lstautogobble}

\title{\Large \textbf{Project-Based Learning for Robot Control Theory: \\ A Robot Operating System (ROS) Based Approach}}
\author{\normalsize Siavash Farzan\\
\normalsize sfarzan@wpi.edu\\
\normalsize Robotics Engineering Department \\
\normalsize Worcester Polytechnic Institute}
\date{}

\makeatletter
\patchcmd{\@maketitle}{\begin{center}}{\begin{adjustwidth}{0.5in}{0.5in}\begin{center}}{}{}
\patchcmd{\@maketitle}{\end{center}}{\end{center}\end{adjustwidth}}{}{}
\makeatother

\begin{document}
\raggedright
\maketitle
\thispagestyle{empty}

\section*{Abstract}

Control theory is an important cornerstone of the robotics field and is considered a fundamental subject in an undergraduate and postgraduate robotics curriculum. Furthermore, project-based learning has shown significant benefits in engineering domains, specifically in interdisciplinary fields such as robotics which require hands-on experience to master the discipline adequately. However, designing a project-based learning experience to teach control theory in a hands-on setting can be challenging, due to the rigor of mathematical concepts involved in the subject. Moreover, access to reliable hardware required for a robotics control lab, including the robots, sensors, interfaces, and measurement instruments, may not be feasible in developing countries and even many academic institutions in the US. The current paper presents a set of six project-based assignments for an advanced postgraduate Robot Control course. The sequence of assignments naturally builds on each other, and by the end of the course, students will be able to develop an advanced control framework in a real-world robotic simulation setting. The assignments leverage the Robot Operating System (ROS), an open-source set of tools, libraries, and software, which is a de facto standard for the development of robotics applications. The use of ROS, along with its physics engine simulation framework, Gazebo, provides a hands-on robotics experience equivalent to working with real hardware. The topics incorporated in the assignments include a set of linear and nonlinear control concepts, including dynamics modeling and analysis of multibody robotic systems, Jacobian linearization and characterization of equilibria in nonlinear state-space models, stabilization via state-feedback control, state estimation and control via observer design, linear quadratic regulator (LQR) control, trajectory tracking using inverse dynamics and feedback linearization, Lyapunov-based robust control design, and the formulation of adaptive controllers. The control design formulations are carried out in MATLAB, while the resulting controllers are applied to the robot in Gazebo through the ROS interface, which can be implemented in C++, Python, or MATLAB using the MATLAB ROS toolbox. Learning outcomes include: i) theoretical analysis of linear and nonlinear dynamical systems, ii) formulation and implementation of advanced model-based robot control algorithms using classical and modern control theory, and iii) programming and performance evaluation of robotic systems on physics engine robot simulators. Course evaluations and student surveys demonstrate that the proposed project-based assignments successfully bridge the gap between theory and practice, and facilitate learning of control theory concepts and state-of-the-art robotics techniques through a hands-on approach.\looseness=-1

\newpage
\section{Introduction}\label{sec:intro}

Control theory is a key foundation in the fields of robotics and engineering and is an essential subject in both undergraduate and postgraduate engineering curricula.
It provides a mathematical framework for analyzing and designing systems that operate autonomously or under the guidance of a human operator, and enables engineers to design control systems that regulate the behavior of a machine or robot to achieve a desired output. Such control systems are ubiquitous in modern times and can be found in a variety of domains, including robotics, aerospace, automotive, chemical, electrical, and biomedical engineering.

When it comes to teaching engineering subjects, project-based learning (PBL) has shown significant benefits~\cite{larmer2015setting,wobbe2019project,larmer2015gold1}, as it allows students to gain hands-on experience in applying the concepts they have learned in class, reinforces their understanding of the concepts, and prepares them for real-world engineering applications. PBL is particularly effective in interdisciplinary fields such as robotics~\cite{Qidwai11}, in which a combination of knowledge and skills from various disciplines, including electrical engineering, mechanical engineering, computer science, and mathematics is required~\cite{Omar2019}. PBL allows students to integrate this knowledge and develop skills such as communication, project management, and problem-solving, which are essential for success in their future careers.

However, designing a project-based learning experience to teach control theory is a challenging task due to the rigorous mathematical concepts involved in the subject~\cite{Omahony2008}. Those concepts are fundamental to understanding and designing control systems, and a PBL framework should strike a balance between rigorous mathematical concepts and hands-on experience to be effective.
To achieve this balance, the projects should incorporate a combination of mathematical models and simulations to help students understand and analyze the behavior and performance of control systems.
In addition, designing a PBL experience that is both engaging and challenging while covering the rigorous mathematical concepts is also essential to motivate and inspire students~\cite{larmer2015gold2}. The projects should be designed to challenge students to apply their understanding of control theory to real-world problems.\looseness=-1

This hands-on control experience for PBL can be achieved through the use of cyber-physical systems, such as robots or process control systems, to allow students to apply the mathematical concepts in practical applications. 
Despite the potential benefits of hands-on experience in engineering education, there may be challenges in providing access to reliable hardware, particularly in developing countries and even some academic institutions in the US. The cost of robotics hardware, sensors, interfaces, and measurement instruments can be prohibitive, especially for institutions that lack adequate funding.
Moreover, even when institutions have access to the necessary hardware, maintaining it and keeping it up-to-date can be challenging.
This can lead to a lack of access to the latest technology and resources, limiting the quality of education available to students.\looseness=-1

It is important to note that the lack of access to reliable hardware and equipment can lead to disparities in engineering education opportunities, particularly for students from underprivileged backgrounds. This can perpetuate social inequality and limit opportunities for these students to enter careers in engineering and robotics.
To address these challenges, institutions can explore alternative approaches to robotics education that do not require expensive hardware, such as simulation software and technological resources. Such approaches can provide students with the necessary knowledge and skills to enter the field of robotics while minimizing the cost of hardware.
Exploring alternative approaches and ensuring that students from all backgrounds have access to quality education in robotics can help to address these challenges and promote equality in the field.\looseness=-1

Considering the opportunities and challenges described above, this paper presents a PBL framework for an advanced postgraduate -- or an upper-level undergraduate -- \textit{Robot Control} course. The course comprehensively covers control theory for robotic systems, including both linear and nonlinear control.

The topics included in the course outline are as follows:
\begin{itemize}\setlength\itemsep{0pt}
    \item[--] Mathematical modeling of dynamical systems
    \item[--] State-space representation of linear and nonlinear systems
    \item[--] Stability analysis and Lyapunov stability
    \item[--] Jacobian linearization
    \item[--] State-feedback control and eigenvalue assignment
    \item[--] State estimation and observer design
    \item[--] Stabilization through output feedback
    \item[--] Linear quadratic regulator (LQR) control
    \item[--] Trajectory generation and tracking for robotic systems
    \item[--] Feedback linearization control
    \item[--] Lyapunov-based robust and adaptive inverse dynamics control
    \item[--] Sliding mode control (SMC)
    \item[--] Force and impedance control
    \item[--] Control Lyapunov functions (CLFs)
    \item[--] Model predictive control (MPC)
    \item[--] Control of robotic manipulators and mobile robots
    through MATLAB and Robot Operating System (ROS)
\end{itemize}

The course learning outcomes defined for students are listed below.
\begin{itemize}
    \item[1.] Develop an understanding of the fundamental principles of robotic systems control and linear/nonlinear control theory.
    \item[2.] Formulate and implement model-based control algorithms for robotic systems that achieve specified control objectives.
    \item[3.] Develop proficiency in using programming languages and software tools for implementing control algorithms in robotic systems.
    \item[4.] Analyze the performance of control algorithms for robotic systems in terms of stability, robustness, optimality, and accuracy.
    \item[5.] Develop critical thinking and problem-solving skills by troubleshooting and improving the performance of control algorithms for robotic systems.
    \item[6.] Develop communication skills by presenting project results and findings to peers and instructors.
    \item[7.] Gain practical experience in working with robotics platforms in physics engine robot simulators.\looseness=-1
\end{itemize}

In light of the opportunities presented by project-based learning and the challenges of providing access to reliable hardware for robotics education, we have developed a sequence of six project-based assignments that provide students with a hands-on experience equivalent to working with actual hardware. These assignments have been designed to align with a project-based teaching philosophy.
The sequence of assignments builds on each other, gradually introducing more complex concepts and allowing students to develop advanced control frameworks with different objectives in a real-world robotic simulation setting. 
The assignments leverage the Robot Operating System (ROS)~\cite{ros}, an open-source set of tools, libraries, and software that is widely used in the development of robotics applications. The use of ROS, along with its physics engine simulation framework, Gazebo~\cite{gazebo}, provides students with a hands-on robotics experience equivalent to working with real hardware.
Utilizing Gazebo as a simulator enables students to design and evaluate their robot control algorithms in ROS. One benefit of this is that the codes and programs developed by students can be readily transferred to the actual robot hardware.

Physics-engine based robot simulators, such as Gazebo, provide an effective and powerful tool for learning and experimenting with robotics systems. They offer several advantages over traditional simulation frameworks like MATLAB in the context of project-based learning and hands-on experience, such as providing realistic motions, greater flexibility, and safety.
Students can easily modify the robot's parameters, such as mass, shape, and size, to see how they affect the robot's behavior.
More importantly, the Gazebo framework offers a comparable experience to working with real hardware. 
Since it uses a physics engine to simulate the behavior of objects in the environment, students can design and evaluate their control algorithms in a realistic simulation environment that models real-world physics, sensor readings, and actuator responses, and closely approximates the behavior of the actual robot hardware.
Gazebo provides real-time sensor data, allowing students to test and refine their robot control algorithms using realistic sensor data. This helps to ensure that the algorithms are robust and reliable in the face of real-world sensor noise and other environmental factors. At the same time, it provides a safe environment for students to experiment and learn without the risk of damaging physical equipment or injuring themselves. This allows students to focus on the development and testing of their projects without the fear of costly mistakes.

The rest of the paper is organized as follows:
Section 2 introduces RRBot -- a two-link robot arm used for the project-based assignments -- and details the set up of its necessary toolchain in ROS and Gazebo.
Section 3 provides an overview of the objectives and technical content covered in the six project-based assignments, along with the results and performance attained in each assignment.
Section 4 offers a comprehensive analysis of the student learning outcomes through the presentation of performance metrics and evaluations, aimed at assessing the efficacy of the proposed project-based learning framework.
Section 5 concludes by reflecting on the outcomes of the proposed project-based learning approach, offering insights into future directions and potential improvements, and discussing approaches for developing other engineering courses using a similar methodology.\looseness=-1

\section{Setting Up RRBot in ROS and Gazebo}
In this section, we outline Assignment 0, designed to assist students in setting up a reliable toolchain for the proposed PBL framework.
For the project-based assignments in the course, we use the RRBot model, which stands for ``Revolute-Revolute Manipulator Robot''. The RRBot model is commonly used as a starting point for more complex robot models, as it provides a relatively simple but realistic example of a robot with joint constraints and dynamics.

\begin{figure}[!ht]
    \centering
    \begin{tabular}{cc}
    \includegraphics[trim={90pt 90pt 40pt 50pt},clip, width=0.43\textwidth]{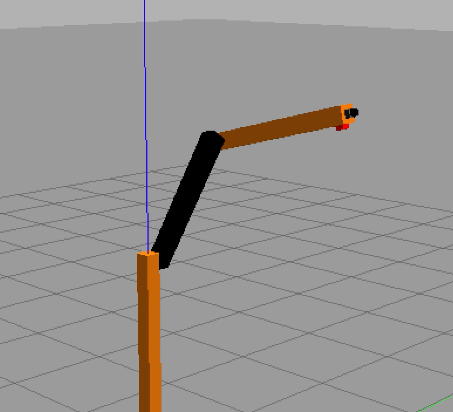} & 
    \centering

\tikzset{every picture/.style={line width=0.75pt}}     

\begin{tikzpicture}[x=0.75pt,y=0.75pt,yscale=-1.6,xscale=1.6]

\draw  [fill={rgb, 255:red, 155; green, 155; blue, 155 }  ,fill opacity=1 ] (1.07,125.83) .. controls (1.07,122.61) and (3.68,120) .. (6.9,120) -- (6.9,120) .. controls (10.12,120) and (12.74,122.61) .. (12.74,125.83) -- (12.74,137.33) .. controls (12.74,137.33) and (12.74,137.33) .. (12.74,137.33) -- (1.07,137.33) .. controls (1.07,137.33) and (1.07,137.33) .. (1.07,137.33) -- cycle ;
\draw   (58.32,55.5) .. controls (57.22,52.96) and (58.38,50.02) .. (60.91,48.91) -- (143.42,12.98) .. controls (145.96,11.88) and (148.9,13.04) .. (150,15.57) -- (150,15.57) .. controls (151.11,18.1) and (149.95,21.05) .. (147.42,22.15) -- (64.9,58.08) .. controls (62.37,59.19) and (59.42,58.03) .. (58.32,55.5) -- cycle ;
\draw   (3.67,129.04) .. controls (1.49,127.35) and (1.1,124.2) .. (2.79,122.02) -- (58,50.95) .. controls (59.7,48.77) and (62.84,48.37) .. (65.02,50.07) -- (65.02,50.07) .. controls (67.2,51.76) and (67.59,54.9) .. (65.9,57.08) -- (10.69,128.16) .. controls (8.99,130.34) and (5.85,130.73) .. (3.67,129.04) -- cycle ;
\begin{scope}[radius=0.28em]
\fill (31.1,93.67) -- ++(0.28em,0) arc [start angle=0,end angle=90] -- ++(0,-0.56em) arc [start angle=270, end angle=180];%
\draw (31.1,93.67) circle;%
\end{scope}
\begin{scope}[radius=0.28em]
\fill (100.1,37.2) -- ++(0.28em,0) arc [start angle=0,end angle=90] -- ++(0,-0.56em) arc [start angle=270, end angle=180];%
\draw (100.1,37.2) circle;%
\end{scope}
\draw [color={rgb, 255:red, 0; green, 37; blue, 255 }  ,draw opacity=1 ] [dash pattern={on 4.5pt off 4.5pt}]  (14.9,132) -- (40.19,101.5) ;
\draw [shift={(40.19,101.5)}, rotate = 129.67] [color={rgb, 255:red, 0; green, 37; blue, 255 }  ,draw opacity=1 ][line width=0.75]    (0,3.35) -- (0,-3.35)(6.56,-1.97) .. controls (4.17,-0.84) and (1.99,-0.18) .. (0,0) .. controls (1.99,0.18) and (4.17,0.84) .. (6.56,1.97)   ;
\draw [shift={(14.9,132)}, rotate = 309.67] [color={rgb, 255:red, 0; green, 37; blue, 255 }  ,draw opacity=1 ][line width=0.75]    (0,3.35) -- (0,-3.35)(6.56,-1.97) .. controls (4.17,-0.84) and (1.99,-0.18) .. (0,0) .. controls (1.99,0.18) and (4.17,0.84) .. (6.56,1.97)   ;
\draw [color={rgb, 255:red, 0; green, 0; blue, 255 }  ,draw opacity=1 ] [dash pattern={on 4.5pt off 4.5pt}]  (67.24,64) -- (103.74,48.33) ;
\draw [shift={(103.74,48.33)}, rotate = 156.77] [color={rgb, 255:red, 0; green, 0; blue, 255 }  ,draw opacity=1 ][line width=0.75]    (0,3.35) -- (0,-3.35)(6.56,-1.97) .. controls (4.17,-0.84) and (1.99,-0.18) .. (0,0) .. controls (1.99,0.18) and (4.17,0.84) .. (6.56,1.97)   ;
\draw [shift={(67.24,64)}, rotate = 336.77] [color={rgb, 255:red, 0; green, 0; blue, 255 }  ,draw opacity=1 ][line width=0.75]    (0,3.35) -- (0,-3.35)(6.56,-1.97) .. controls (4.17,-0.84) and (1.99,-0.18) .. (0,0) .. controls (1.99,0.18) and (4.17,0.84) .. (6.56,1.97)   ;
\draw [color={rgb, 255:red, 150; green, 150; blue, 150 }  ,draw opacity=1 ] [dash pattern={on 4.5pt off 4.5pt}]  (62.74,53) -- (102.07,3) ;
\draw  [draw opacity=0] (7.07,98.75) .. controls (7.78,98.69) and (8.51,98.67) .. (9.24,98.67) .. controls (14.75,98.67) and (19.84,100.25) .. (23.91,102.91) -- (9.24,119.33) -- cycle ; \draw  [color={rgb, 255:red, 255; green, 0; blue, 0 }  ,draw opacity=1 ] (7.07,98.75) .. controls (7.78,98.69) and (8.51,98.67) .. (9.24,98.67) .. controls (14.75,98.67) and (19.84,100.25) .. (23.91,102.91) ;
\draw  [draw opacity=0] (82.1,27.25) .. controls (83.54,28.63) and (84.87,30.2) .. (86.04,31.96) .. controls (87.96,34.84) and (89.27,37.92) .. (89.99,41.02) -- (68.85,43.43) -- cycle ; \draw  [color={rgb, 255:red, 255; green, 0; blue, 0 }  ,draw opacity=1 ] (82.1,27.25) .. controls (83.54,28.63) and (84.87,30.2) .. (86.04,31.96) .. controls (87.96,34.84) and (89.27,37.92) .. (89.99,41.02) ;
\draw [color={rgb, 255:red, 255; green, 0; blue, 0 }  ,draw opacity=1 ] [dash pattern={on 4.5pt off 4.5pt}]  (62.74,53) -- (99.74,37.67) ;
\draw [color={rgb, 255:red, 255; green, 0; blue, 0 }  ,draw opacity=1 ] [dash pattern={on 4.5pt off 4.5pt}]  (7.24,124.5) -- (31.07,93.67) ;
\draw   (4.74,124.5) .. controls (4.74,123.12) and (5.86,122) .. (7.24,122) .. controls (8.62,122) and (9.74,123.12) .. (9.74,124.5) .. controls (9.74,125.88) and (8.62,127) .. (7.24,127) .. controls (5.86,127) and (4.74,125.88) .. (4.74,124.5) -- cycle ;
\draw  [draw opacity=0] (33.73,68.31) .. controls (38.53,69.27) and (43.16,71.43) .. (47.17,74.82) .. controls (51.14,78.18) and (54.02,82.35) .. (55.77,86.88) -- (27.79,97.72) -- cycle ; \draw  [color={rgb, 255:red, 144; green, 19; blue, 254 }  ,draw opacity=1 ] (33.73,68.31) .. controls (38.53,69.27) and (43.16,71.43) .. (47.17,74.82) .. controls (51.14,78.18) and (54.02,82.35) .. (55.77,86.88) ;
\draw  [color={rgb, 255:red, 144; green, 19; blue, 254 }  ,draw opacity=1 ] (56.18,81.02) .. controls (55.49,83.11) and (55.36,84.88) .. (55.79,86.36) .. controls (54.79,85.18) and (53.24,84.32) .. (51.12,83.74) ;
\draw  [draw opacity=0] (118.46,15.96) .. controls (121.89,19.46) and (124.5,23.85) .. (125.91,28.91) .. controls (127.3,33.92) and (127.33,38.98) .. (126.23,43.71) -- (97,36.93) -- cycle ; \draw  [color={rgb, 255:red, 144; green, 19; blue, 254 }  ,draw opacity=1 ] (118.46,15.96) .. controls (121.89,19.46) and (124.5,23.85) .. (125.91,28.91) .. controls (127.3,33.92) and (127.33,38.98) .. (126.23,43.71) ;
\draw  [color={rgb, 255:red, 144; green, 19; blue, 254 }  ,draw opacity=1 ] (129.56,40.41) .. controls (127.82,41.76) and (126.71,43.15) .. (126.24,44.62) .. controls (126.07,43.09) and (125.27,41.5) .. (123.83,39.84) ;
\draw  [color={rgb, 255:red, 255; green, 0; blue, 0 }  ,draw opacity=1 ] (21.2,98.17) .. controls (21.59,100.34) and (22.33,101.96) .. (23.42,103.04) .. controls (21.99,102.49) and (20.21,102.48) .. (18.07,103) ;
\draw  [color={rgb, 255:red, 255; green, 0; blue, 0 }  ,draw opacity=1 ] (91.58,36.46) .. controls (90.68,38.46) and (90.37,40.23) .. (90.65,41.74) .. controls (89.78,40.47) and (88.31,39.44) .. (86.26,38.67) ;
\draw [color={rgb, 255:red, 150; green, 150; blue, 150 }  ,draw opacity=1 ] [dash pattern={on 4.5pt off 4.5pt}]  (7.24,124.5) -- (7.19,5) ;

\draw (30.07,115.4) node [anchor=north west][inner sep=0.75pt]  [font=\Large,color={rgb, 255:red, 0; green, 0; blue, 255 }  ,opacity=1 ]  {$r_{1}$};
\draw (84.07,58.4) node [anchor=north west][inner sep=0.75pt]  [font=\Large,color={rgb, 255:red, 0; green, 0; blue, 255 }  ,opacity=1 ]  {$r_{2}$};
\draw (11.1,83.4) node [anchor=north west][inner sep=0.75pt]  [font=\Large,color={rgb, 255:red, 255; green, 0; blue, 0 }  ,opacity=1 ]  {$\theta_{1}$};
\draw (87.0,20.4) node [anchor=north west][inner sep=0.75pt]  [font=\Large,color={rgb, 255:red, 255; green, 0; blue, 0 }  ,opacity=1 ]  {$\theta_{2}$};
\draw (40,91) node [anchor=north west][inner sep=0.75pt]  [font=\Large]  {$m_{1}$};
\draw (106.07,40.4) node [anchor=north west][inner sep=0.75pt]  [font=\Large]  {$m_{2}$};
\draw (59.07,77.4) node [anchor=north west][inner sep=0.75pt]  [font=\Large,color={rgb, 255:red, 144; green, 19; blue, 254 }  ,opacity=1 ]  {$\tau_{1}$};
\draw (131.07,32.4) node [anchor=north west][inner sep=0.75pt]  [font=\Large,color={rgb, 255:red, 144; green, 19; blue, 254 }  ,opacity=1 ]  {$\tau_{2}$};

\end{tikzpicture}
    \end{tabular}
    \caption{The 2-DoF revolute-revolute robot arm (RRBot).}
    \label{fig:rrbot}
\end{figure}
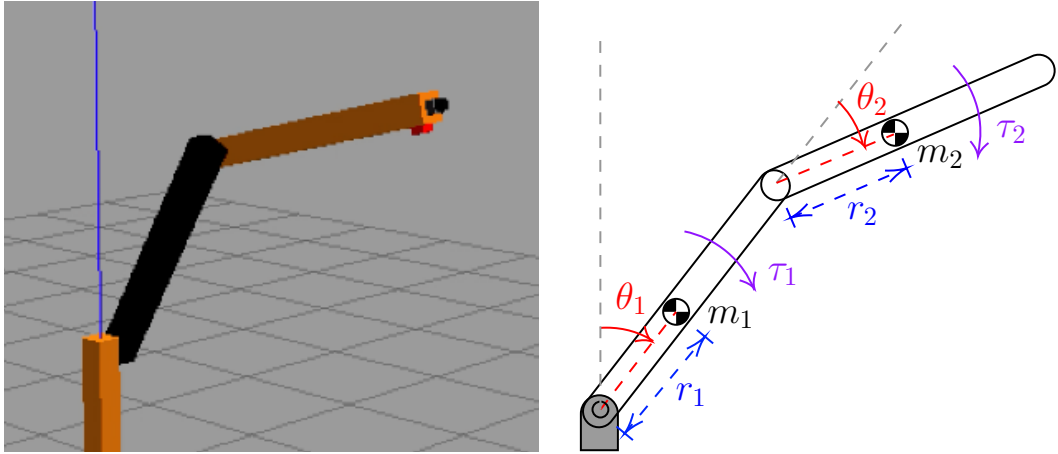

The robot, as shown in Figure~\ref{fig:rrbot}, consists of two revolute joints and three linkages. The first link is connected to a fixed pole via a joint, while the second link can rotate around the second joint. 
Below are the physical parameters of the robot, including the mass ($m_1,\,m_2$), length ($l_1,\,l_2$), center of mass location ($r_1,\,r_2$), and inertia ($I_1,\,I_2$) of the first and second links:
\begin{gather*}
    m_1=m_2=1\; \textrm{(kg)}, \quad l_1=l_2=1\; \textrm{(m)}, \quad r_1=r_2=0.45\; \textrm{(m)}, \quad
    I_1=I_2=0.084\; \textrm{(kg}\cdot \textrm{m}^2\textrm{)} 
\end{gather*}

The robot's generalized coordinates can be defined as $\theta_1$ and $\theta_2$, which respectively represent the angle between the first link and the vertical and the angle between the first and second link. The robot is equipped with two actuators that apply torque inputs, $\tau_1$ and $\tau_2$, to the respective joints, as illustrated in Figure~\ref{fig:rrbot}.

The RRBot model is available in Gazebo as an example robot model~\cite{rrbot-get}, which can be used to develop and test robot control algorithms in ROS. This open-source model is accessible through a public repository~\cite{rrbot-git,sfarzan-git} and can be easily loaded into Gazebo using the appropriate configuration files. By leveraging this model, researchers and students can experiment with and improve robot control algorithms without requiring access to physical robots, enabling a more cost-effective and efficient development process.

An updated version of the RRBot repository, featuring a ROS launch file for joint effort controllers, can be found in the accompanying GitHub repository for this paper, accessible at: \\
\url{https://github.com/sfarzan/pbl_robot_control}

To ensure a productive learning experience for students throughout the course, it is essential to establish a reliable toolchain setup at the outset. By doing so, students can focus on the technical aspects of the assignments, unencumbered by issues related to dependencies in Linux or missing packages for ROS. This approach promotes a more efficient and effective learning environment, where students can confidently and quickly engage with and progress through the course assignments.\looseness=-1

To facilitate this, Assignment 0 -- which must be completed in the first week of the course before beginning Assignment 1 -- provides students with instructions to install the necessary operating system and software, including Linux Ubuntu 20.04~\cite{linux-ubuntu}, Robot Operating System (ROS) Noetic~\cite{ros-noetic}, and MATLAB~\cite{matlab} with relevant toolboxes including ROS Toolbox~\cite{matlab-ros} and Symbolic Math Toolbox~\cite{matlab-smt}.
Additionally, students are directed to download the RRBot Gazebo package for ROS by cloning the RRBot repository~\cite{sfarzan-git}, and install the required ROS and Python dependencies, specifically \texttt{python3-rosdep}, \texttt{python3-rosinstall} and \texttt{build-essential}.

\begin{figure}[!ht]
    \centering
    \includegraphics[width=\textwidth]{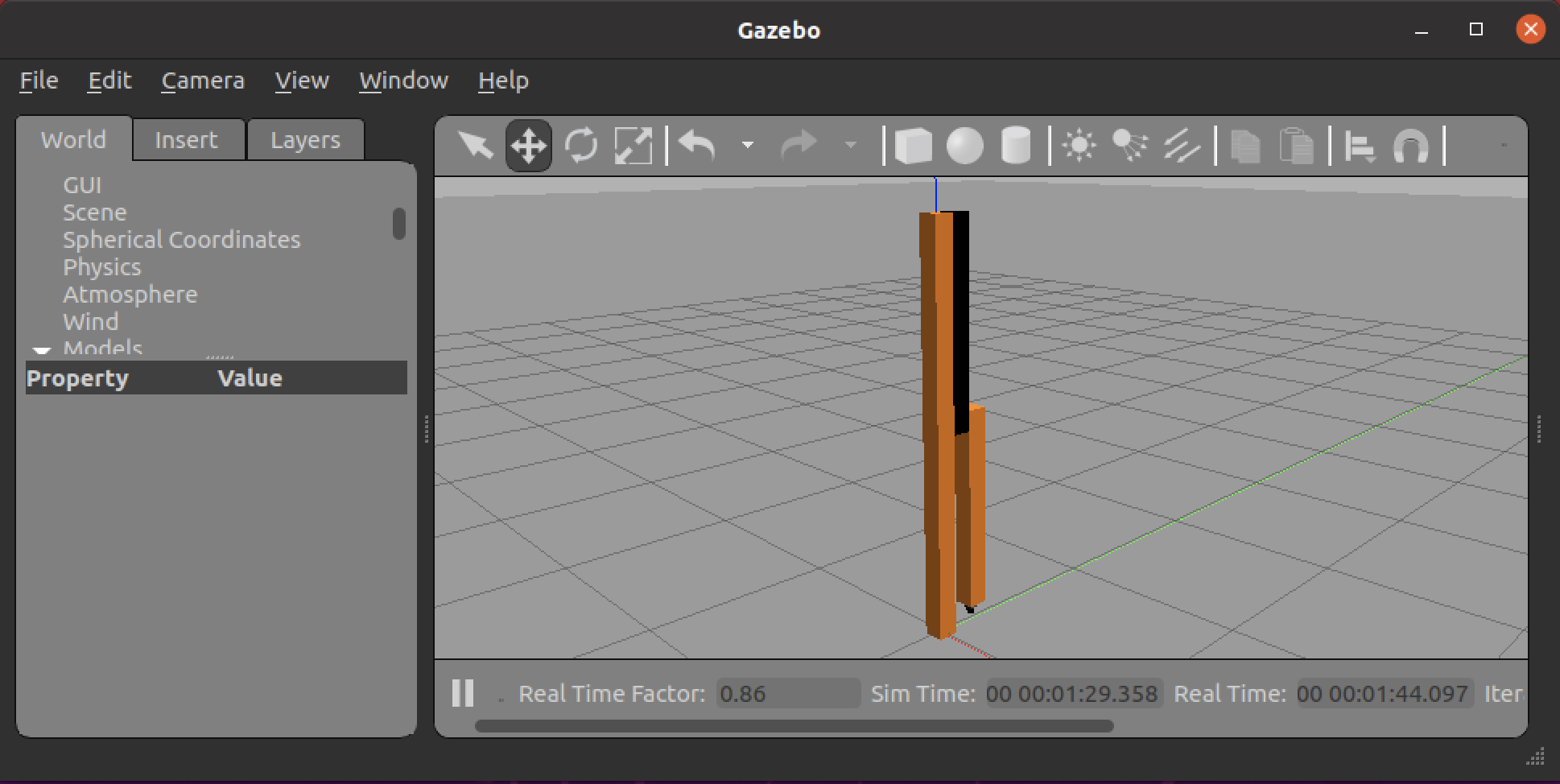}
    \caption{The RRBot model, displayed in its resting position, within the Gazebo environment.}
    \label{fig:rrbot-gazebo}
\end{figure}

To verify that the installation steps have been successfully completed, students should be able to launch and visualize the RRBot model in Gazebo. This can be achieved by running the following command in the Ubuntu terminal, which will launch the robot in Gazebo, as depicted in Figure~\ref{fig:rrbot-gazebo}.
\begin{lstlisting}
    roslaunch rrbot_gazebo rrbot_world.launch
\end{lstlisting}

By verifying that the RRBot model is visible in Gazebo, students can confirm that the installation has been successful, and that they have the necessary tools and resources to begin working on the course assignments.

The use of the RRBot platform in the assignments allows students to apply the control theory concepts they learn in a practical setting, and to see firsthand how the concepts taught in lectures can be used to design controllers that can stabilize and control the motion of the robot.
Gazebo's physics-based simulator makes it easy to work with RRBot's actuators and sensors in ROS, providing students with a hands-on robotics experience that is equivalent to working with real hardware.

The following section presents the project-based assignments designed for the Robot Control course using the RRBot platform.

\section{Assignments}
This section describes the six project-based assignments designed for the Robot Control course. The assignments are intended to provide students with practical experience in the application of control theory to real-world robotic systems. All assignments will use MATLAB, Python, and ROS along with the Gazebo simulation environment. By working with these tools, students will gain hands-on experience that is similar to what they would encounter when working with an actual robot. The assignments will cover a range of topics, including dynamics modeling and analysis of multibody robotic systems, Jacobian linearization and characterization of equilibria in nonlinear state-space models, stabilization via state-feedback control, state estimation and control via observer design, linear quadratic regulator (LQR) control, trajectory tracking using inverse dynamics and feedback-linearization control, Lyapunov-based robust control design, and the formulation of Lyapunov-based adaptive controllers.

\subsection*{Assignment 1: Dynamics Modeling and Analysis of Multi-Body Robotic Systems}
The first project-based assignment focuses on the essential topics of dynamics modeling using the \textit{Lagrangian} approach, state space representation of dynamical systems, and simulation of the two-link robot in MATLAB. The assignment provides students with a practical experience in applying these concepts to robotic systems. They are required to derive the mathematical model (i.e. the equations of motion) of the given two-link robot using the Lagrangian approach, and then obtain the state-space representation of the system. All the formulations and derivations must be carried out symbolically using the Symbolic Math Toolbox~\cite{matlab-smt} in MATLAB. They must also program a simulation of the robot in MATLAB to validate the accuracy of the derived model. The simulation results are then compared to the expected behavior of the system in a real-world scenario in Gazebo, which reinforces the theoretical concepts learned in the lectures. By completing this assignment, students gain practical experience in the modeling and simulation of robotic systems and develop their programming skills using MATLAB.

The Lagrangian method~\cite{murray1994mathematical} is developed by defining the Lagrangian function $L$ as the difference between the robot's kinetic energy $K$ and potential energy $P$, which are functions of the robot's configuration $q=\begin{bmatrix} \theta_1, & \theta_2 \end{bmatrix}^T$ and velocities $\dot{q}=\begin{bmatrix} \dot{\theta}_1, & \dot{\theta}_2 \end{bmatrix}^T$:
\begin{equation}
    L(q,\dot{q}) =K(q,\dot{q})-P(q)
\end{equation}
To derive the Euler-Lagrange equations, the partial derivative of the Lagrangian with respect to each of the generalized coordinates $q_i,\;i=1,2$ are calculated,
and the time derivative of the partial derivative of $L$ with respect to the velocities $\dot{q}_i$ are taken.
The resulting expressions are set equal to the generalized force/torque $u_i$ applied to the respective generalized coordinate. This results in the Euler-Lagrange equation for each $q_i$ in the form of
\begin{equation}
    \frac{d}{dt}\Big(\frac{\partial L}{\partial \dot{q}_i}\Big) - \frac{\partial L}{\partial {q}_i} = u_i, \quad i=1,2
\end{equation}
which form the equations of motion for the robotic system.

Next, students are asked to define a state vector $x=\begin{bmatrix} \theta_1, & \theta_2, & \dot{\theta}_1, & \dot{\theta}_2 \end{bmatrix}^T$ for the robot, and use the \texttt{solve()} function in MATLAB to rewrite the dynamics in \textit{nonlinear state-space form} of $\dot{x}=f(x,u)$, with $u$ defined as the generalized inputs to the system , i.e. $u = \begin{bmatrix} \tau_1, \quad \tau_2 \end{bmatrix}^T$.
    
At the end of the assignments, students are asked to construct a passive simulation (i.e., input torques $u$ set to zero) of the robot in MATLAB using the \texttt{ode45()} function. This function is used to simulate the motion of the robot by integrating the equations of motion from a set of given initial conditions $x(0) = \begin{bmatrix} 30^\circ, \,45^\circ, 0,\,0\end{bmatrix}$ for a finite time span of 10 seconds. Students are then required to plot the resulting trajectories and explain the chaotic behavior of the robot. This experience emphasizes the importance of designing feedback controllers in the next assignments, as it intuitively highlights the need for control mechanisms to stabilize the robot's motion.

Additionally, students are required to run the robot from the same initial conditions in the ROS robotic simulator, Gazebo. They are then required to sample the joints data (i.e. positions and velocities) through either the ROS MATLAB toolbox or a Python script for ROS and plot the data in MATLAB. This task allows students to compare the motion trajectories in MATLAB and Gazebo, and comment on any discrepancies between the two. This comparison helps students understand the effects of real-world dynamics and noise on the robot's motion and the importance of testing control algorithms in a simulated environment before implementing them on the physical robot.\looseness=-1

\begin{figure}[!ht]
    \centering
    \includegraphics[width=\textwidth]{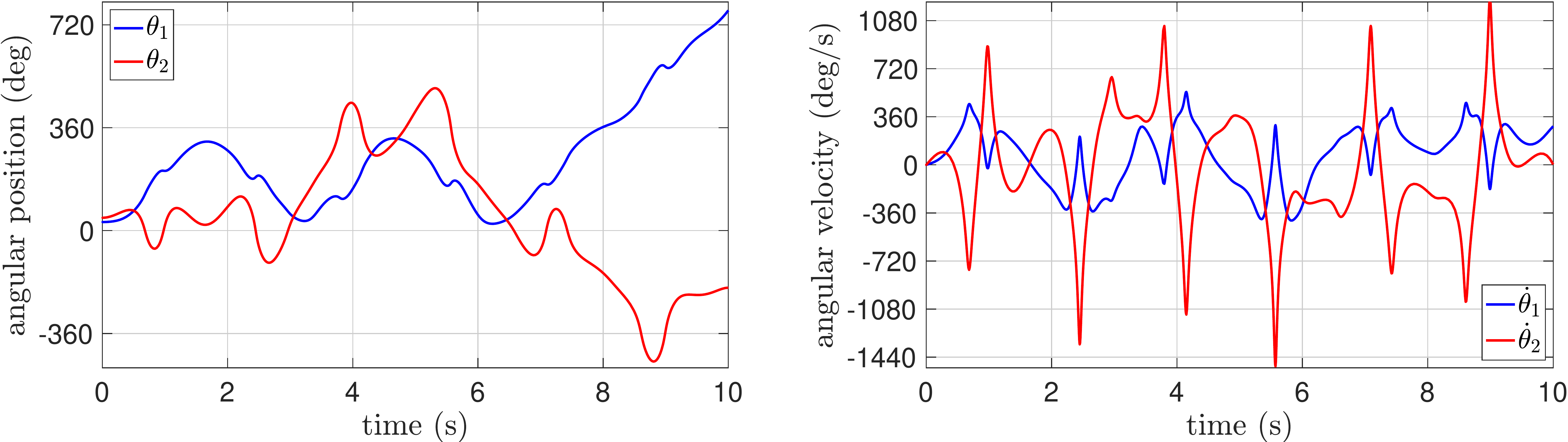}
    \caption{Passive motion trajectories of the robot simulated in MATLAB using the Lagrangian equations of motion for the dynamic model.}
    \label{fig:assignment_1a}
\end{figure}

\begin{figure}[!ht]
    \centering
    \includegraphics[width=\textwidth]{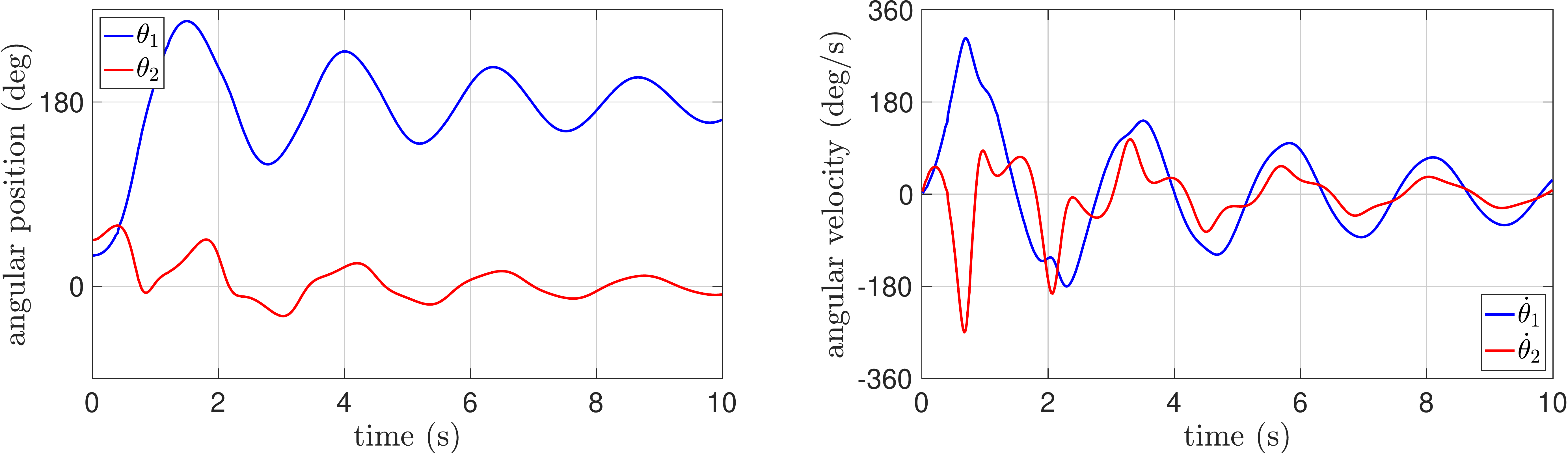}
    \caption{Passive motion trajectories of the actual robot obtained in the Gazebo environment.}
    \label{fig:assignment_1b}
\end{figure}

An example of the motion trajectories obtained from the simulation in MATLAB and the motion of the robot in Gazebo are illustrated in Figures~\ref{fig:assignment_1a} and \ref{fig:assignment_1b}, respectively. The figures provide a visual representation of the motion of the robot, enabling the students to compare and contrast the results obtained from both the simulation and the Gazebo environment. The plots illustrate that the motion trajectories in Gazebo are different from the motions simulated in MATLAB. Such differences can be attributed to the simplifications involved in the mathematical derivations, as well as the presence of frictions and other real-world factors that are not taken into account during the modeling process. The comparison of simulation results with the experimental ones in Gazebo is a valuable learning experience for the students, as it provides them with an understanding of the limitations of simulation and highlights the importance of considering the real-world aspects of the system when designing control algorithms. This exercise also prepares them for the upcoming assignments, where they will design and implement feedback controllers to control the robot in the Gazebo environment.

The resulting equations of motion derived from the Lagrangian method will be used in the next assignments to design various control algorithms for the robot and control its motion in Gazebo.\looseness=-1

\subsection*{Assignment 2: Stabilization via State-Feedback Control and Linear Quadratic Regulation (LQR) Control }
In the second assignment for Robot Control, students are tasked with designing a state-feedback controller to stabilize a two-link robot at its upward equilibrium point. The first step in this process is to rewrite the symbolic equations of motion derived in the previous assignment in \textit{nonlinear state-space form}, and plug in the physical parameters of the robot to obtain numerical equations. Using MATLAB, students then determine the \textit{equilibrium points} by finding the solutions to $f(x,0)=0$, and \textit{linearize} the dynamics around each equilibrium
to write the system dynamics in the linear state-space form of
\vspace{-3pt}
\begin{equation}
    \dot{x}=Ax+Bu, \quad A\in\mathbb{R}^{4\times 4},\; B\in\mathbb{R}^{4 \times 2}
\end{equation}
The \textit{stability} properties of equilibrium points are analyzed by investigating the eigenvalues of the linearized state matrix $A$, gaining a comprehensive understanding of the robot's behavior and control.

Having the linearized system around the upward equilibrium, the students then investigate the \textit{controllability} of the system and design a \textit{state-feedback controller} for the robot using the eigenvalue assignment method~\cite{hespanha2018linear} by deriving a control law in the form of
\vspace{-3pt}
\begin{equation}
    u=-Kx, \quad K\in\mathbb{R}^{2 \times 4}
\end{equation}
such that the closed-loop dynamics in Eq.~\eqref{eq:ass2-dyn} have eigenvalues at some selected desired locations.
\vspace{-3pt}
\begin{equation}\label{eq:ass2-dyn}
    \dot{x} = \big(A-BK\big)x
\end{equation}
The eigenvalues of the closed-loop system must be tuned such that the controller converges to its upward equilibrium point within 10 seconds, while maintaining the control inputs within a the bounds of $-20 \leq \tau_1 \leq 20$ Nm and $ -10 \leq \tau_2 \leq 10$ Nm.

Building upon the design of the state-feedback controller, the students then design the controller's gains $K$ using the \textit{Linear Quadratic Regulator (LQR)} control law~\cite{Farzan2019}:
\vspace{-3pt}
\begin{equation}
    K=R^{-1}B^TP
\end{equation}
where matrix $P=P^T>0$ is the solution to the Algebraic Riccati Equation:
\vspace{-3pt}
\begin{equation}
    A^TP + PA + Q - PBR^{-1}B^TP = 0.
\end{equation}
The weighting matrices $Q\geq0$ and $R>0$ (with $Q\in\mathbb{R}^{4\times4}$ and $R\in\mathbb{R}^{2 \times 2}$) must be selected and tuned to satisfy the same performance requirements specified for the state-feedback controller designed earlier.

To verify the performance of the controllers, students simulate the robot under the designed control laws in MATLAB (starting from the initial conditions of $\theta_1(0)=30^{\circ}$, $\theta_2(0)=45^{\circ}$, $\dot{\theta}_1(0)=0$ and $\dot{\theta}_2(0)=0$), and verify if the robot converges to the upward equilibrium within the specified time span of 10 seconds while satisfying the control bounds. If the performance is not satisfactory, students need to go back to the control design and change the state-feedback controller's eigenvalues or the weighting matrices in the LQR formulation and simulate the system again. Once the controllers performance are satisfactory, students implement the control laws in their ROS Python framework to control the robot in Gazebo.

At the conclusion of the assignment, students are required to include the plots of the state trajectories as well as the control inputs to evaluate and compare the performance of both the state-feedback and LQR controllers implemented in MATLAB and Gazebo environments. A detailed analysis of the performance of the robot is expected, discussing the strengths and weaknesses of each controller.
Moreover, it is crucial for students to pinpoint and explain any inconsistencies between the simulated results in MATLAB and those derived from controlling the robot within the Gazebo environment. By addressing these discrepancies, students will demonstrate their understanding of potential real-world challenges and limitations that may arise when transitioning from simulation to practical application. 

\begin{wrapfigure}[25]{R}{0.39\textwidth}
  \begin{center}
    \includegraphics[trim={0cm 0.1cm 0.3cm 0cm},clip, width=0.38\columnwidth]{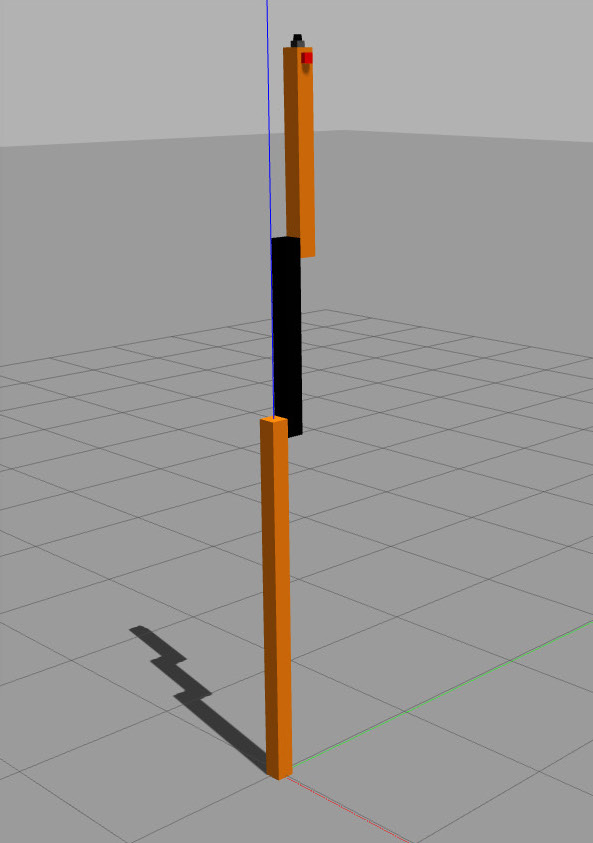}
  \end{center}
  \caption{The two-link robot in Gazebo stabilized to its upward configuration under the state-feedback controller.}
  \label{fig:assignment_2a}
\end{wrapfigure}

\begin{figure}[!ht]
    \centering
    \includegraphics[width=\textwidth]{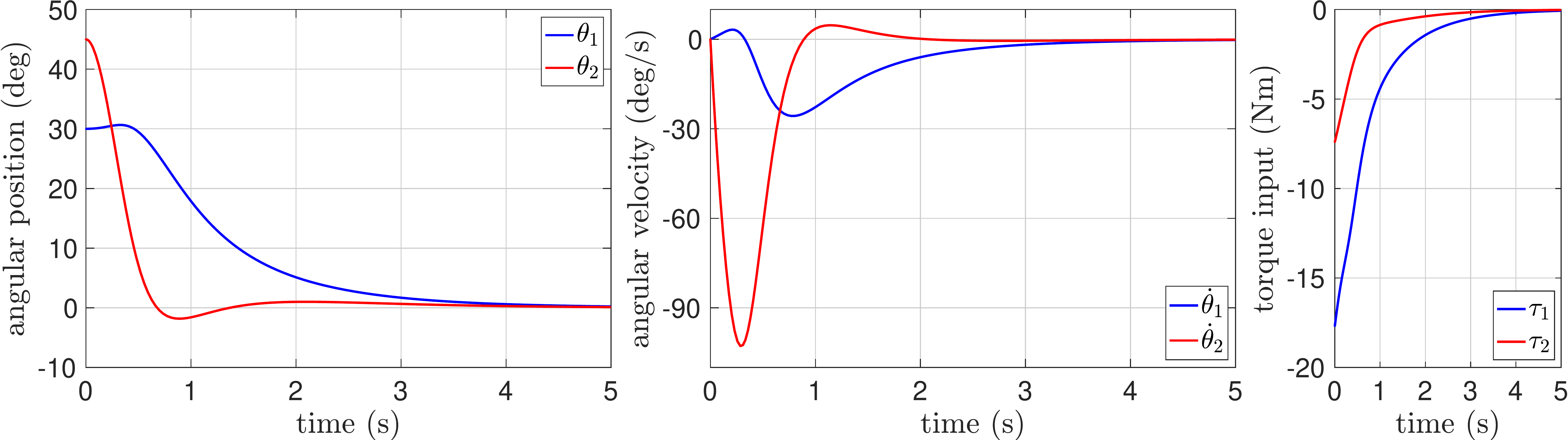}
    \caption{Motion trajectories and control inputs of the robot under the state-feedback control, stabilizing to its upward configuration in approximately four seconds.}
    \label{fig:assignment_2b}
\end{figure}

\begin{figure}[!ht]
    \centering
    \includegraphics[width=\textwidth]{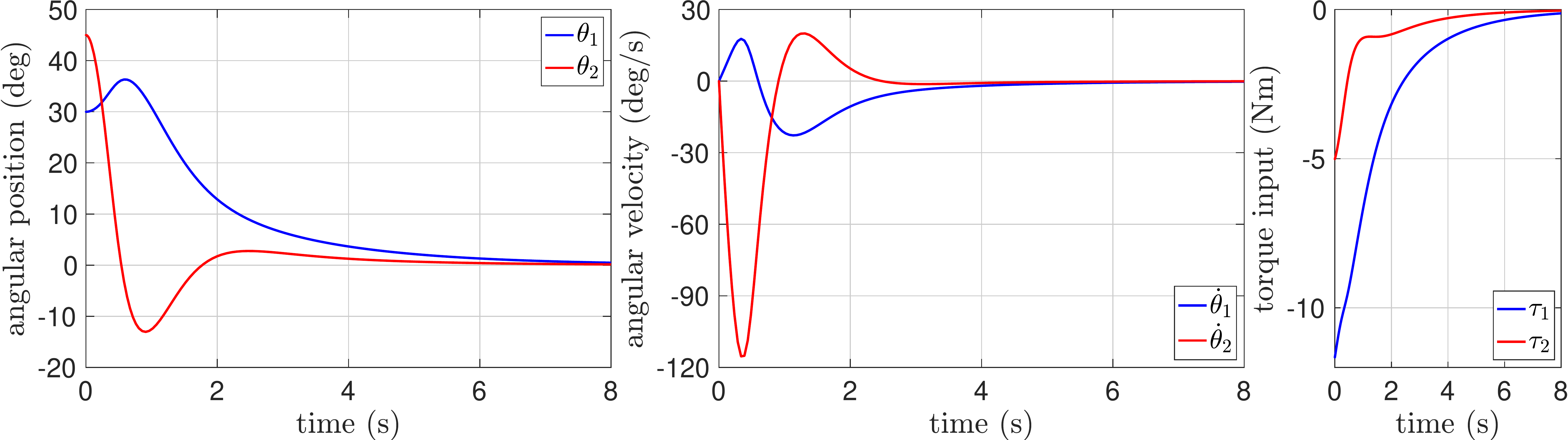}
    \caption{Motion trajectories and control inputs of the robot under the LQR control, stabilizing to its upward configuration in approximately six seconds.}
    \label{fig:assignment_2c}
\end{figure}

The performance of the robot under the state-feedback control law in the Gazebo environment is illustrated in Figure~\ref{fig:assignment_2a}.
It can be seen that the controller has successfully stabilized the robot to its equilibrium point corresponding to its upward configuration.
To provide a visual representation of the motion and control effort of the robot under both the state-feedback and LQR controllers, Figures~\ref{fig:assignment_2b} and \ref{fig:assignment_2c} present an example of the resulting motion and control effort trajectories in both the MATLAB simulation and the Gazebo environment, demonstrating that the system states has all converged to zero, associated with the upward configuration. These figures allow for a direct comparison of the performance of the robot under both controllers, both in simulation and in the real-world. The motion trajectories provide insights into the stability and accuracy of the robot's motion, while the control effort trajectories give an indication of the effort required to control the robot and the optimality of the controller. For example, the control inputs of the robot under the LQR controller in Figure~\ref{fig:assignment_2c} are more optimal when compared to the state-feedback controller shown in Figure~\ref{fig:assignment_2b}. This comparison allows the students to gain a deeper understanding of the trade-offs between performance and optimality in control design, and how to analyze and compare the performance of the robot in simulation and a real-world environment.

\vskip 5pt

\subsection*{Assignment 3: Stabilization using State Estimation and Observer Design}
Assignment 3 focuses on output feedback control through \textit{observer} design for the same two-link robot. Unlike the previous assignment, it is assumed that only the position variables $q=\begin{bmatrix} \theta_1, & \theta_2 \end{bmatrix}$ are measured via the sensors, and not the velocities $\dot{q}=\begin{bmatrix} \dot{\theta}_1, & \dot{\theta}_2 \end{bmatrix}$. To address this issue, a full-state observer~\cite{astrom2021feedback} is designed according to Eq.~(\ref{eq:observer}) to estimate the state of the system $\hat{x}$:
\begin{equation}\label{eq:observer}
    \dot{\hat{x}} = A\hat{x}+Bu + L(y-C\hat{x})
\end{equation}
with the output matrix $C\in\mathbb{R}^{2\times 4}$ defined as $C=\begin{bmatrix} 1 & 0 & 0 & 0 \\ 0 & 1 & 0 & 0 \end{bmatrix}$, and the gains $L\in\mathbb{R}^{4 \times 2}$ computed by rendering the state matrix $A-LC$ associated with the closed-loop error dynamics asymptotically stable.

The estimated states $\hat{x}$ are used in real-time in the same state-feedback control law designed in Assignment 2, as shown in Eq.~(\ref{eq:xhat-law}):
\begin{equation}\label{eq:xhat-law}
    u=-K\hat{x}, \quad K\in\mathbb{R}^{2 \times 4}
\end{equation}
The convergence and stability of the overall system with the observer is guaranteed according to the \textit{separation principle}~\cite{antsaklis2006linear}.

\vspace{-5pt}
\begin{figure}[!hb]
    \centering
    \includegraphics[width=\textwidth]{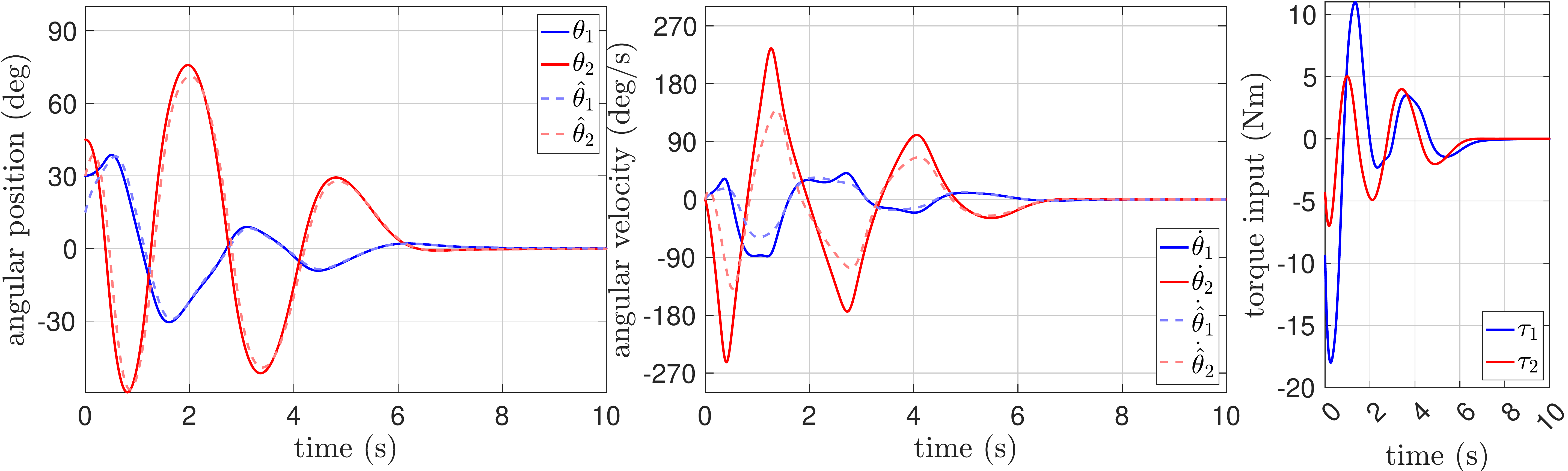}
    \caption{Motion trajectories and control inputs of the robot under the output-feedback control, with estimated velocities converging to actual values and stabilization to its upward configuration in about eight seconds. A full-state observer is employed for state estimation.}
    \label{fig:assignment_3a}
\end{figure}

The performance of the robot under the output feedback control through observer design is evaluated in both MATLAB and Gazebo environments. The robot starts from the same off-nominal initial conditions in Assignments 1.
The plots and the performance of the robot, as shown in Figure~\ref{fig:assignment_3a}, demonstrate that the performance of the output feedback control with estimated states from the observer is comparable to the performance of the nominal controller in Assignment 2, despite estimating the states and using them in the control law, rather than the actual states. The plots also show the convergence of the estimated states to their actual values during the motion. This highlights the effectiveness of observer design in estimating the state variables when not all the states are measurable.
Students gain practical experience in observer design and state estimation as applied in robotics through this assignment.

\subsection*{Assignment 4: Trajectory Tracking via Feedback Linearization}
The fourth project-based assignment for Robot Control focuses on designing a nonlinear feedback-linearization controller for the two-link robot to track a reference trajectory. Students begin by generating dynamically feasible trajectories $q_d$ as \textit{cubic polynomial trajectories}~\cite{spong2020robot,siciliano2010robotics} in the form of
\begin{equation}
    q_d(t) = a_0 + a_1 t + a_2 t^2 + a_3 t^3
\end{equation}
for the two joints of the robot, with $a_i,\,i=0,1,2,3$ as the coefficients of the polynomials to be determined.
The time span and the desired initial and final configuration and velocity of the robot are given by:
$t_f = 10$ seconds, $\theta_1(0) = 180^\circ$, $\theta_1(t_f)=0$, $\theta_2(0) = 90^\circ$, $\theta_2(t_f)=0$, $\dot{\theta}_1(0)=0$, $\dot{\theta}_1(t_f)=0$, $\dot{\theta}_2(0)=0$, and $\dot{\theta}_2(t_f)=0$.

This is followed by rewriting the second-order equations of motion derived in the first assignment in the standard manipulator form:
\vspace{-3pt}
\begin{equation}
    M(q)\ddot{q} + C(q,\dot{q})\dot{q} + g(q) = u 
\end{equation}
in which $M(q)\in\mathbb{R}^{2\times 2}$ is the symmetric positive definite inertial matrix, $C(q,\dot{q})\in\mathbb{R}^{2\times2}$ represents the Coriolis and centripetal forces, and the vector $g(q)\in\mathbb{R}^{2}$ contains the gravitational forces.

Using the equations of motion in the manipulator form, the students then derive a symbolic \textit{feedback linearization controller}~\cite{slotine1991applied}, also known as \textit{inverse dynamics control}~\cite{siciliano2010robotics}, to cancel the nonlinearities:
\vspace{-3pt}
\begin{equation}
    u = M(q) v + C(q,\dot{q})\dot{q} + g(q)
\end{equation}
The suggested control law renders the system as a double integrator $\ddot{q}=v$, with the term $v\in\mathbb{R}^2$ as a virtual control input to be designed.

Next, to ensure the linear error dynamics accurately track the reference trajectory, the double integrator system is written in the state-space form:
\vspace{-3pt}
\begin{equation}
    \dot{x}=Ax+Bv \quad \textrm{with} \quad A=\begin{bmatrix} 0 & I \\ 0 & 0 \end{bmatrix}, \; B = \begin{bmatrix} 0 \\ I \end{bmatrix}
\end{equation}
Specifying the error variables $\xi=\begin{bmatrix}e, & \dot{e} \end{bmatrix}^T$, $\xi \in \mathbb{R}^4$, with the error terms defined as $e:=q-q_d$ and $\dot{e}:=\dot{q}-\dot{q}_d$, the state-feedback control method is employed to design the virtual control input as:
\begin{equation}\label{eq:virtual}
    v = -K \xi + \ddot{q}_d, \quad K\in\mathbb{R}^{2\times4}
\end{equation}
The feedback gains $K$ are designed by stabilizing the closed-loop dynamics $\dot{x}=(A-BK)x$ for the double integrator system. The feedforward control term $\ddot{q}_d$ is included in the control law in Eq.~\eqref{eq:virtual} to track the reference trajectory.

At the end, students update their MATLAB simulation framework as well as their Python ROS script for the robot in Gazebo developed in the previous assignments with the feedback linearization control law derived in this assignment. They then evaluate the performance of the system in tracking the reference trajectory to reach the upward position, when the robot starts from off-nominal configuration of
$\theta_1(0) = 200^\circ$, $\theta_2(0) = 125^\circ$, $\dot{\theta}_1(0)=0$, and $\dot{\theta}_2(0)=0$.
If the performance is not satisfactory (i.e. the system does not track and converge to the reference trajectory), the students can adjust and update the state-feedback control gains for the virtual control input until the system converges to the reference trajectory.
The controller should be again tuned such that the control inputs remain within the desired bounds specified in Assignment 2, that is $-20 \leq \tau_1 \leq 20$ Nm and $ -10 \leq \tau_2 \leq 10$ Nm.

\begin{figure}[!ht]
    \centering
    \includegraphics[width=\textwidth]{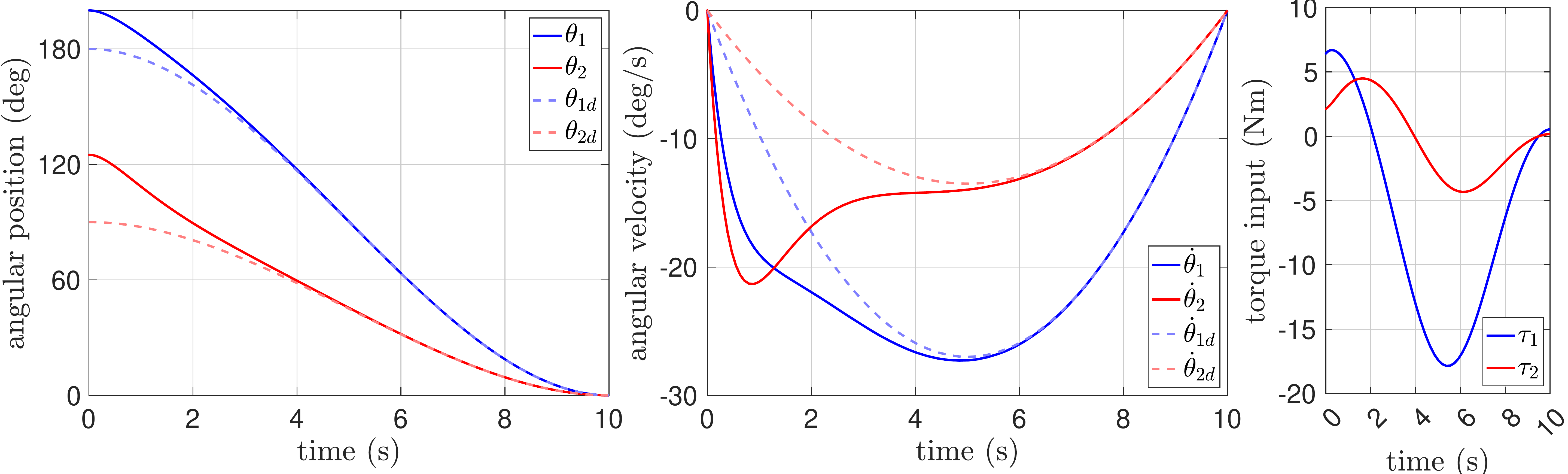}
    \caption{Motion trajectories and control inputs of the robot under the feedback-linearization control, starting from an off-nominal initial configuration and converging to the reference trajectory within approximately five seconds.}
    \label{fig:assignment_4a}
\end{figure}

\begin{figure}[!ht]
\renewcommand{\arraystretch}{0.4}
\setlength\tabcolsep{2pt}
\begin{tabular}{ccc}
{\includegraphics[trim={50cm 10.5cm 26cm 13.9cm},clip, width=0.32\columnwidth]{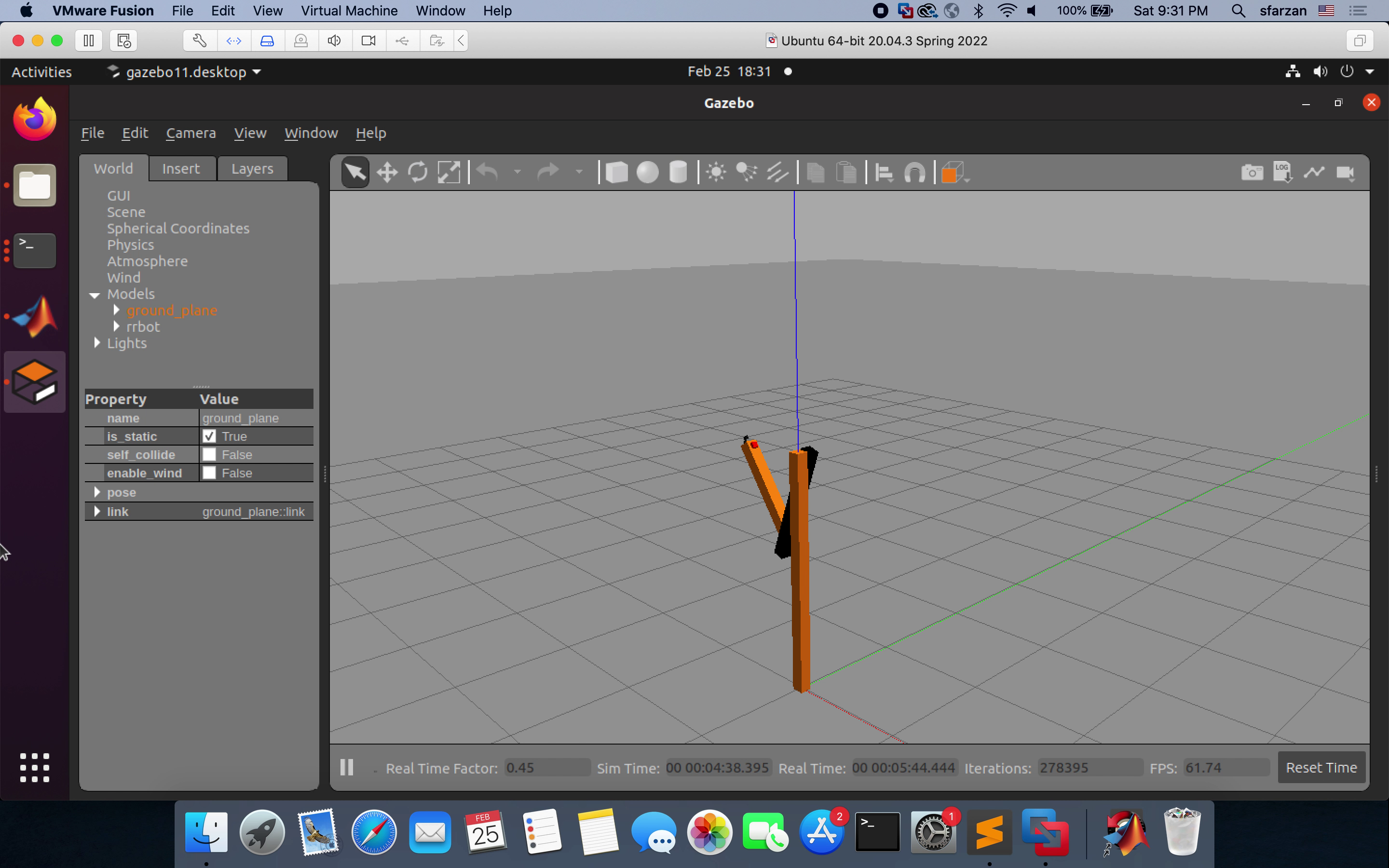}} & 
{\includegraphics[trim={50cm 10.5cm 26cm 13.9cm},clip, width=0.32\columnwidth]{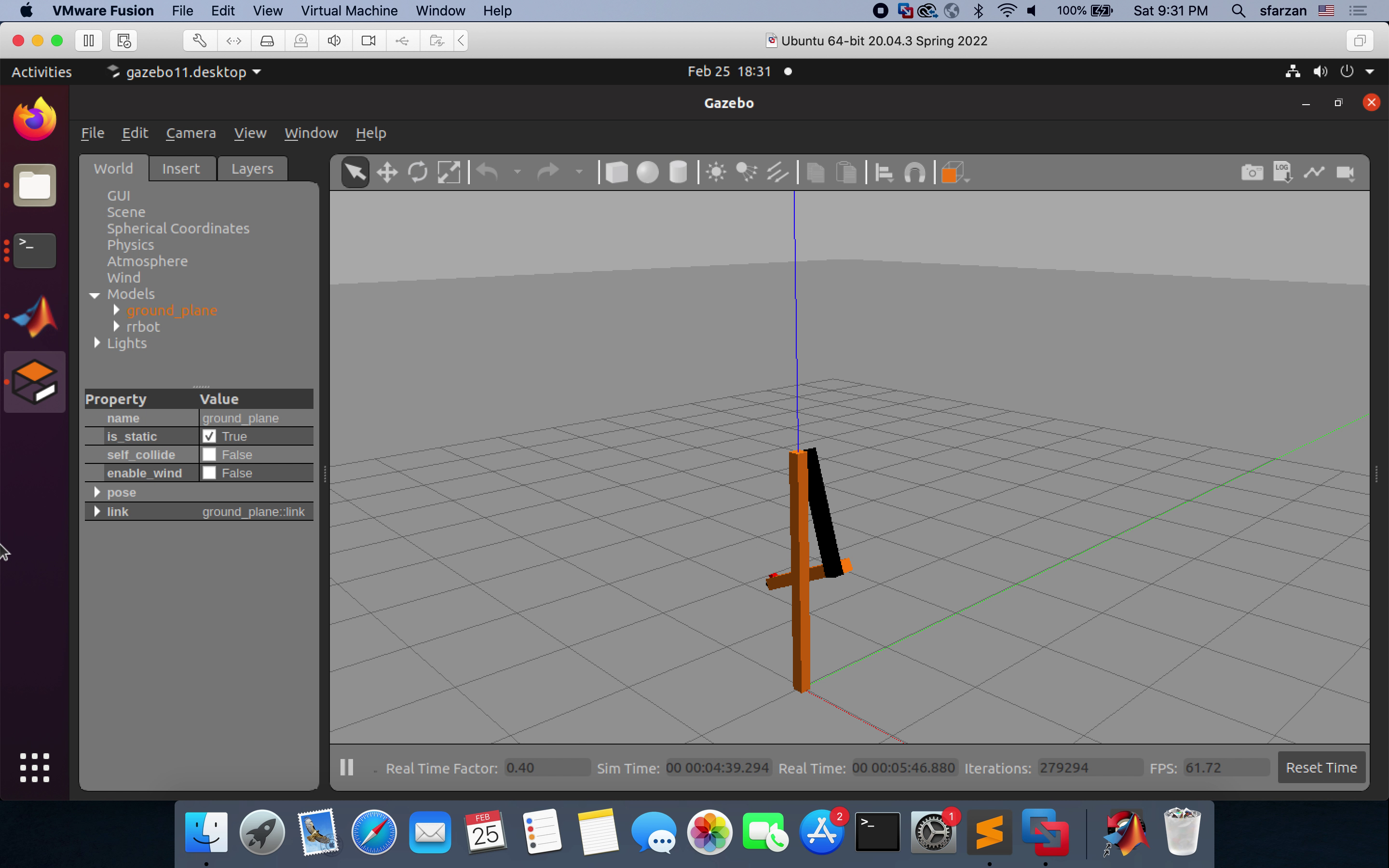}} &
{\includegraphics[trim={50cm 10.5cm 26cm 13.9cm},clip, width=0.32\columnwidth]{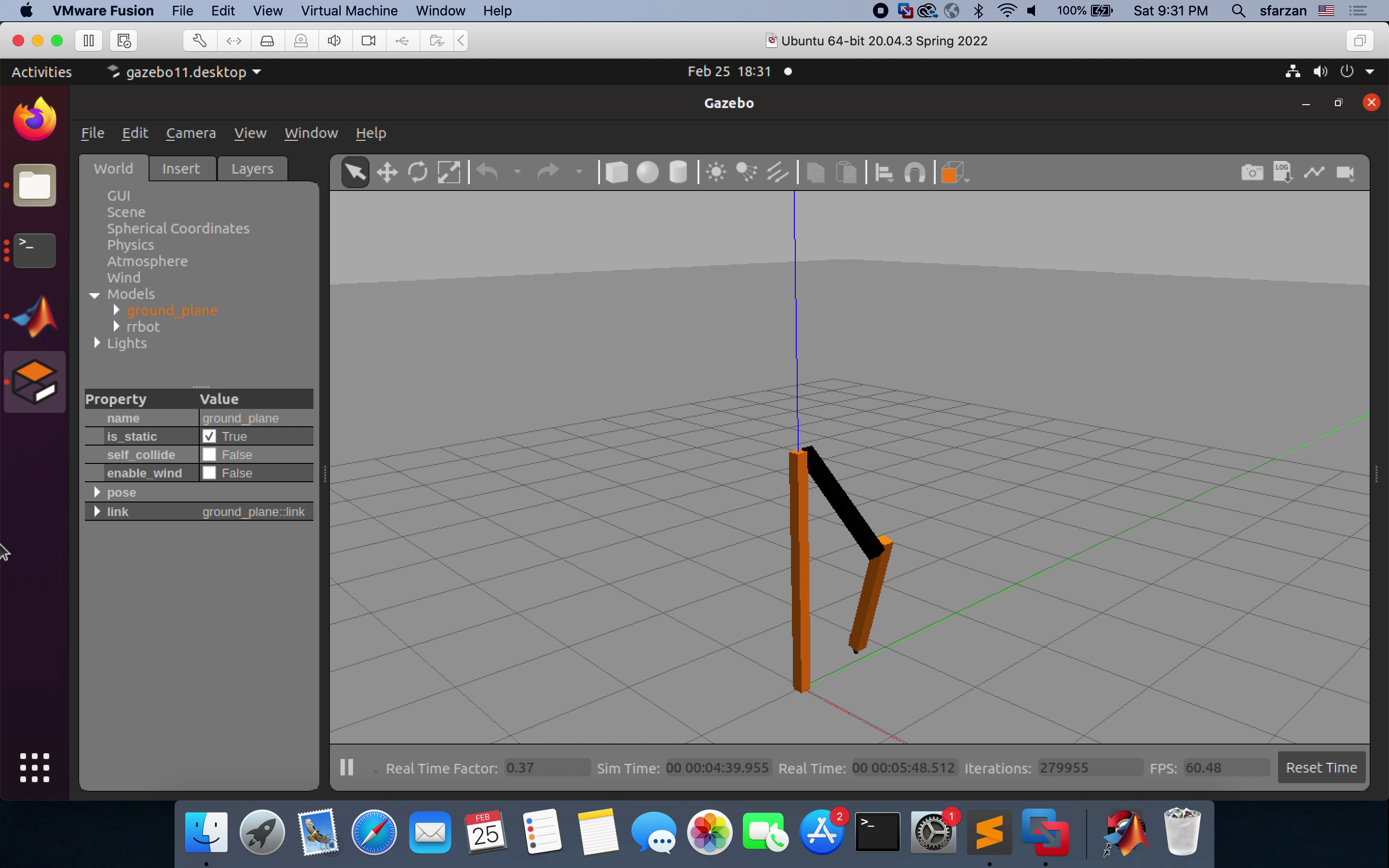}}
\tabularnewline
{(a)} & {(b)} & {(c)} \tabularnewline
{\includegraphics[trim={50cm 10.5cm 26cm 13.9cm},clip, width=0.32\columnwidth]{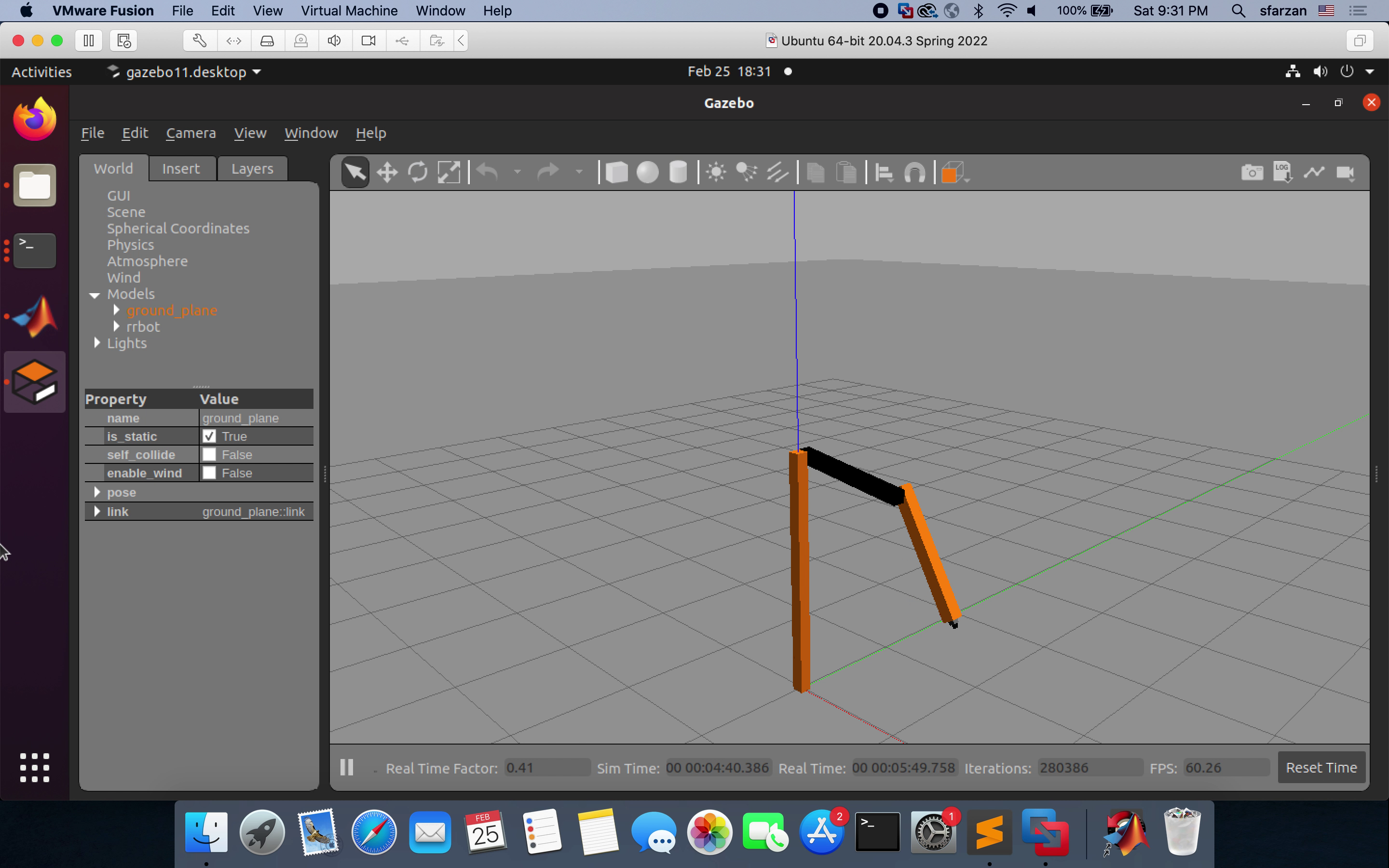}} & 
{\includegraphics[trim={50cm 10.5cm 26cm 13.9cm},clip, width=0.32\columnwidth]{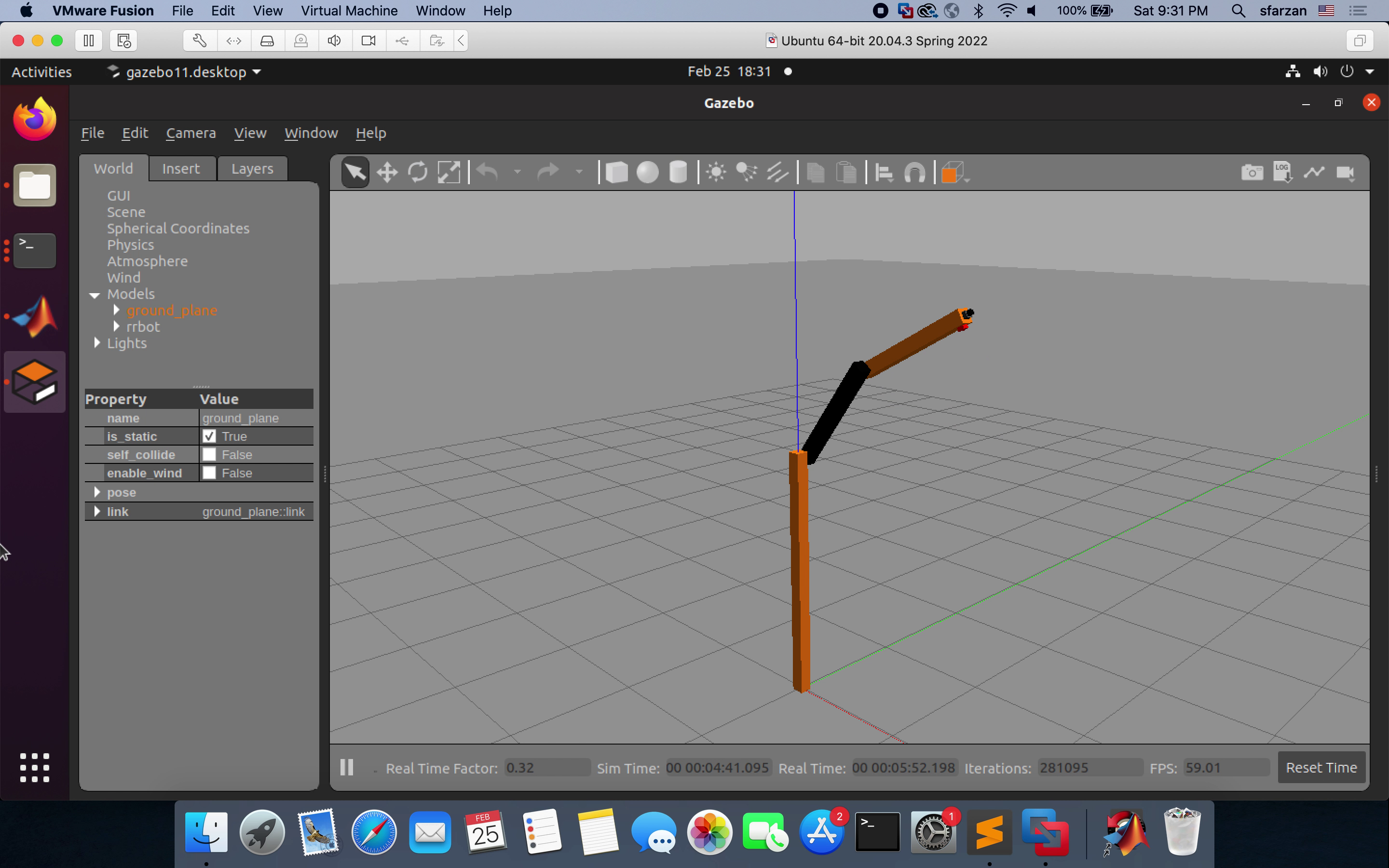}} &
{\includegraphics[trim={50cm 10.5cm 26cm 13.9cm},clip, width=0.32\columnwidth]{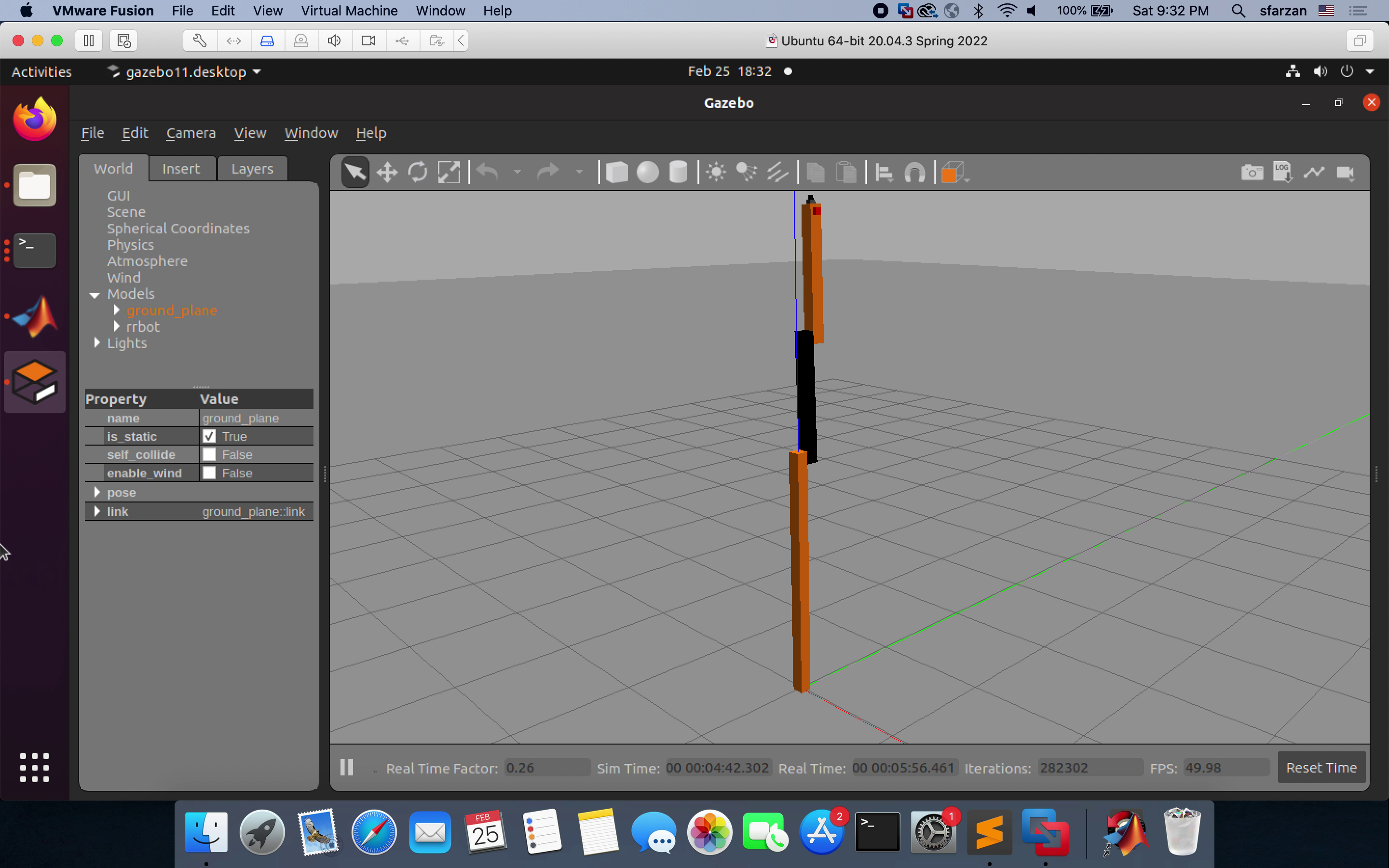}}
\tabularnewline
{(d)} & {(e)} & {(f)} \tabularnewline
\end{tabular}
\caption{\label{fig:assignment_4b} Trajectory tracking by the robot utilizing feedback-linearization control within the Gazebo environment. Starting from an off-nominal initial configuration, the robot successfully converges to the reference trajectory and reaches the desired upward configuration in about ten seconds.\looseness=-1}
\end{figure}

Finally, students plot the state trajectories, the control inputs trajectories, and the associated reference trajectories for both the MATLAB simulation and Gazebo experiment, and analyze the performance of the robot. An example of such performance is demonstrated in Figure~\ref{fig:assignment_4a}, showing the asymptotic convergence of the state trajectories to the reference trajectories while maintaining the control inputs within the actuator bounds.
Figure~\ref{fig:assignment_4b} illustrates the successful trajectory tracking of the robot in Gazebo using the feedback linearization control. Starting from the off-nominal configuration, the robot is able to track the reference trajectory and reach the upward configuration in approximately ten seconds.

This assignment provides students with the opportunity to gain experience with \textit{trajectory tracking}. The assignment builds upon the foundation established in the previous assignments, where students focused on \textit{stabilizing} the system to its upward equilibrium point using linear control techniques.
This assignment not only enhances the students' understanding of the nonlinear control of robotic systems by feedback-linearization, but also gives them a glimpse of the challenges and complexities involved in trajectory tracking, which is a common problem in the field of robotics.

\subsection*{Assignment 5: Lyapunov-based Robust Control Design for Trajectory Tracking under Model Uncertainty}
In the fifth project-based assignment for Robot Control, students build upon the nonlinear feedback linearization controller designed in Assignment 4 to develop a Lyapunov-based \textit{robust inverse dynamics} controller. This controller is used to track the same reference trajectories for the two-link robot, but with the added challenge of dynamic model uncertainties. The robust control design in this assignment takes into consideration the fact that the mass and inertia parameters of the robot (i.e. $m_1$, $m_2$, $I_1$ and $I_2$) are unknown, but nominal values $\hat{m}_1$, $\hat{m}_2$, $\hat{I}_1$ and $\hat{I}_2$ are provided as 75\% of the actual values.

The control law suggested for this assignment is based on the inverse dynamics control designed in Assignment 4, but with nominal values:
\begin{equation}\label{eq:control-law-5}
    u = \hat{M}(q) v + \hat{C}(q,\dot{q})\dot{q} + \hat{g}(q)
\end{equation}

To address the uncertainties in the dynamic model parameters, a robust control term $v_r\in\mathbb{R}^2$ is included in the virtual control law for the inverse dynamics controller designed in Assignment~4:\looseness=-1
\begin{equation}
    v = -K \xi + \ddot{q}_d + v_r, \quad K\in\mathbb{R}^{2\times4}
\end{equation}
The feedback gains $K$ are designed following the same approach as in Assignment 4, with the aim of achieving asymptotic stability and tracking performance in the presence of model uncertainties. The robust control term is designed to ensure that the system remains stable and converges to the desired trajectory even when there are unknown disturbances or uncertainties present.

To design the robust control term $v_r$, students first determine an upper bound term $\rho$ for the uncertainty term in the dynamics. Based on the Lyapunov analysis for robust inverse dynamics derived in the course lectures~\cite{de1996theory}, the robust control term can then be designed using the closed-form solution:
\begin{equation}
    v_r = \begin{cases} -\rho(\xi,t)\,\dfrac{B^TP\xi}{\norm{B^TP\xi}} & \;\textrm{if} \, \norm{B^TP\xi}\neq 0\\
    0 & \;\textrm{if} \,\norm{B^TP\xi}=0\end{cases}
\end{equation}
The matrix $P\in\mathbb{R}^{4 \times 4}$ is obtained as the solution to the Lyapunov equation $A_{cl}^TP+PA_{cl}=-Q$, where $A_{cl}=A-BK$, and $Q\in\mathbb{R}^{4\times 4}$ is a positive definite tuning parameter.

In addition to addressing the uncertainty in the dynamic model, students also need to tackle the chattering effect that occurs in the control input. To accomplish this, a boundary layer $\phi>0$ is included in the robust control term:
\begin{equation}
    v_r = \begin{cases} -\rho(\xi,t)\,\dfrac{B^TP\xi}{\norm{B^TP\xi}} & \;\textrm{if} \,  \norm{B^TP\xi} > \phi\\
    -\rho(\xi,t)\,\dfrac{B^TP\xi}{\phi} & \;\textrm{if} \,  \norm{B^TP\xi} \leq \phi\end{cases}    
\end{equation}
The width of the boundary layer must be tuned to achieve a balance between tracking accuracy and effective chattering removal. This technique allows students to gain hands-on experience with the trade-offs between performance and robustness in control design, and how to design controllers that are both accurate and robust in the presence of uncertainties and model mismatch, a valuable skill in the field of robotics.

Students apply their newly developed Lyapunov-based robust inverse dynamics controller to the two-link robot in the Gazebo environment. To achieve this, they update the MATLAB simulation and the Python ROS script previously developed in the fourth assignment with the new robust control law. This provides students with an opportunity to evaluate the effectiveness of their control design and to observe the trajectory tracking performance of the robot under the presence of model uncertainties. The controller should be again tuned such that the control inputs remain within the desired bounds.
Despite the model uncertainties, the performance of the robot in the Gazebo environment is similar to the performance in Assignment 4 (shown in Figure~\ref{fig:assignment_4b}), demonstrating the effectiveness of the robust control design.

\begin{figure}[!ht]
    \centering
    \includegraphics[width=\textwidth]{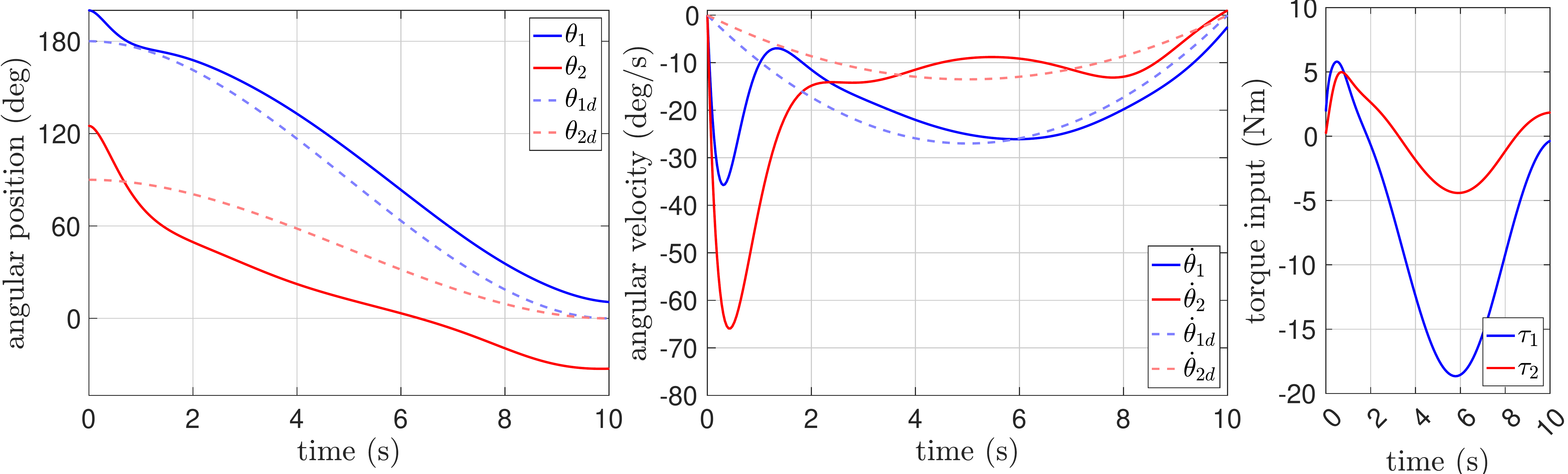}
    \caption{Motion trajectories and control inputs of the robot under the feedback-linearization control with uncertain nominal values (without the robust control term), highlighting the robot's failure to converge to the reference trajectory.}
    \label{fig:assignment_5a}
\end{figure}

\begin{figure}[!ht]
    \centering
    \includegraphics[width=\textwidth]{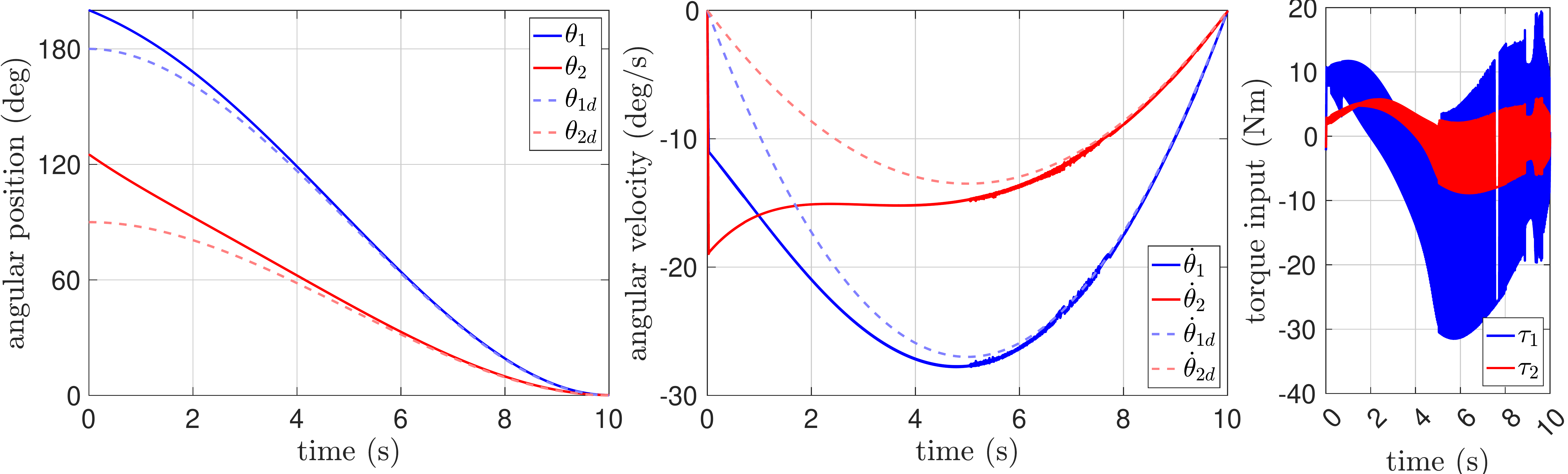}
    \caption{Motion trajectories and control inputs of the robot using robust control without a boundary layer. Despite model uncertainties, the robot converges to the reference trajectory in about seven seconds, albeit with significant chattering in the control inputs.}
    \label{fig:assignment_5b}
\end{figure}

\begin{figure}[!ht]
    \centering
    \includegraphics[width=\textwidth]{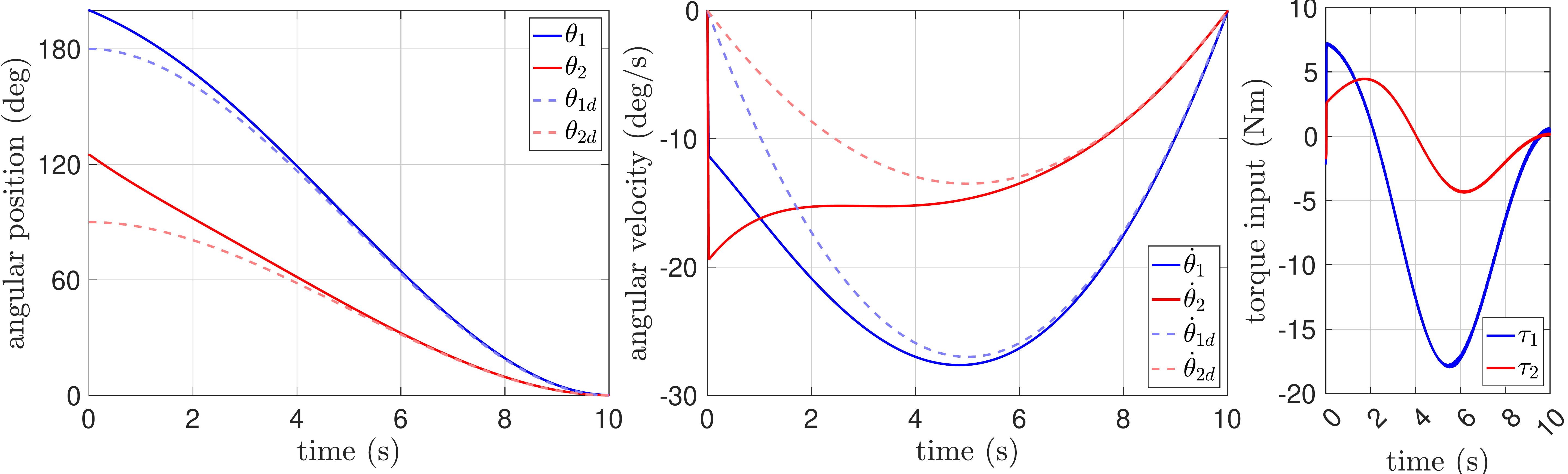}
    \caption{Motion trajectories and control inputs of the robot employing robust control with a boundary layer, demonstrating improved performance by converging to the reference trajectory without chattering in the control inputs.}
    \label{fig:assignment_5c}
\end{figure}

Students are tasked with performing a comprehensive comparison and analysis of three scenarios: controlling the robot with the control design in Assignment 4, but with the nominal values (to establish a baseline for the poor performance of the original controller), controlling the robot using the robust control without boundary layer, and controlling the robot using the robust control with boundary layer. Sample plots for the state trajectories, the control inputs trajectories, and the associated reference trajectories for these scenarios are shown in Figures~\ref{fig:assignment_5a}, \ref{fig:assignment_5b}, and \ref{fig:assignment_5c}, respectively. By comparing these scenarios, students can visually analyze the performance of the controller in terms of tracking accuracy, stability, robustness, and chattering.

\vskip 2pt


\subsection*{Assignment 6: Lyapunov-based Adaptive Control Design for Trajectory Tracking under Model Uncertainty}
In the last project-based assignment for the course, students build upon their progress in the previous assignments and formulate an \textit{adaptive inverse dynamics} control for trajectory tracking of the same two-link robot with model uncertainties. Adaptive control aims to estimate and compensate for the uncertainties present in the robotic system through an online adaptation, thus enhancing the overall control performance.

To that end, the equations of motion derived in Assignment 1 are written in the linear parametric form representing the unknown dynamics of the system as a linear combination of known basis functions with unknown coefficients:
\begin{equation}
    Y(q,\dot{q},\ddot{q})\,\alpha = u 
\end{equation}
In the dynamic equation above, $Y(q,\dot{q},\ddot{q})\in\mathbb{R}^{2\times5}$ is the regressor/basis function, and $\alpha\in\mathbb{R}^{5\times1}$ is the unknown parameter vector, including lumped parameters from the original system.

The objective of this assignment is to design a control law that adapts to the unknown parameters and maintain accurate trajectory tracking performance in the presence of model uncertainties.
The adaptation law, derived in the course lectures~\cite{craig1988adaptive} through a Lyapunov analysis, is then used to update the parameters in real-time:
\begin{equation}
    \dot{\hat{\alpha}} = -\Gamma^{-1}\phi^T B^T P\xi
\end{equation}
with $\Gamma\in\mathbb{R}^{5 \times 5}$ as a symmetric positive definite tuning matrix affecting the parameter adaptation convergence rate, and $\phi=\hat{M}^{-1}(q)Y(q,\dot{q},\ddot{q})$.
As in Assignment 5, the matrix $P$ is obtained by solving the Lyapunov equation $A_{cl}^TP+PA_{cl}=-Q$, with $Q\in\mathbb{R}^{4\times 4}$ as a positive definite tuning parameter.

The adaptation law estimates the unknown parameters $\hat{\alpha}$ to adapt to the dynamic model uncertainties in the system.
The estimated parameters are used in real-time to form the adaptive control law, which is formulated as follows:
\begin{equation}
    u = Y(q,\dot{q},v)\,\hat{\alpha}(t)
\end{equation}
with the same virtual control input $v\in\mathbb{R}^{2}$ designed in Eq.~(\ref{eq:virtual}) for trajectory tracking.

In order to evaluate the performance of the adaptive control law, experiments are conducted using both MATLAB and the Gazebo environment through the ROS Python API. The robot is initialized with the same off-nominal initial conditions as in Assignments 4 and 5, and the initial value for the unknown parameter vector is set to 75\% of the actual parameters, i.e. $\hat{\alpha}(0) = 0.75\alpha$. State and control input trajectories are compared against those generated by the nominal controller with no adaptive terms.
Students are encouraged to further explore the design parameters, such as $K$, $Q$ and $\Gamma$, and tune them accordingly to achieve the desired performance in case the initial performance is not satisfactory. This thorough evaluation allows students to gain experience in analyzing and tuning adaptive control systems for real-world scenarios with model uncertainties.

\begin{figure}[!ht]
    \centering
    \includegraphics[width=\textwidth]{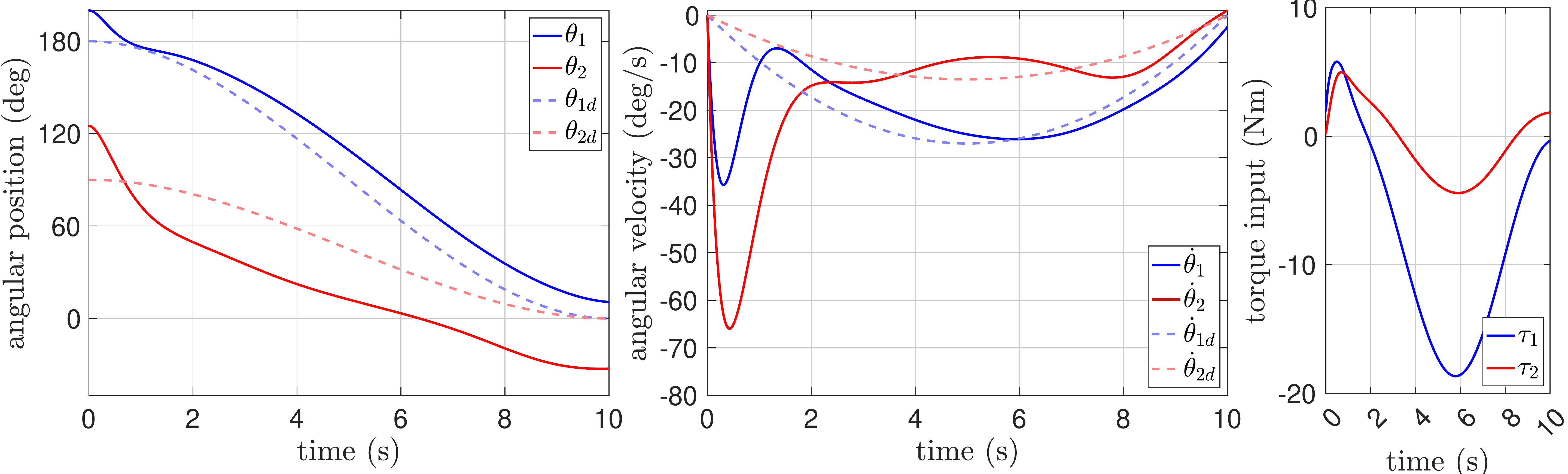}
    \caption{Motion trajectories and control inputs of the robot under the nominal inverse dynamics control using the uncertain parameters $\hat{\alpha}(0)$ (without adaptation), emphasizing the robot's inability to converge to the reference trajectory.}
    \label{fig:assignment_6a}
\end{figure}

\begin{figure}[!ht]
    \centering
    \includegraphics[width=\textwidth]{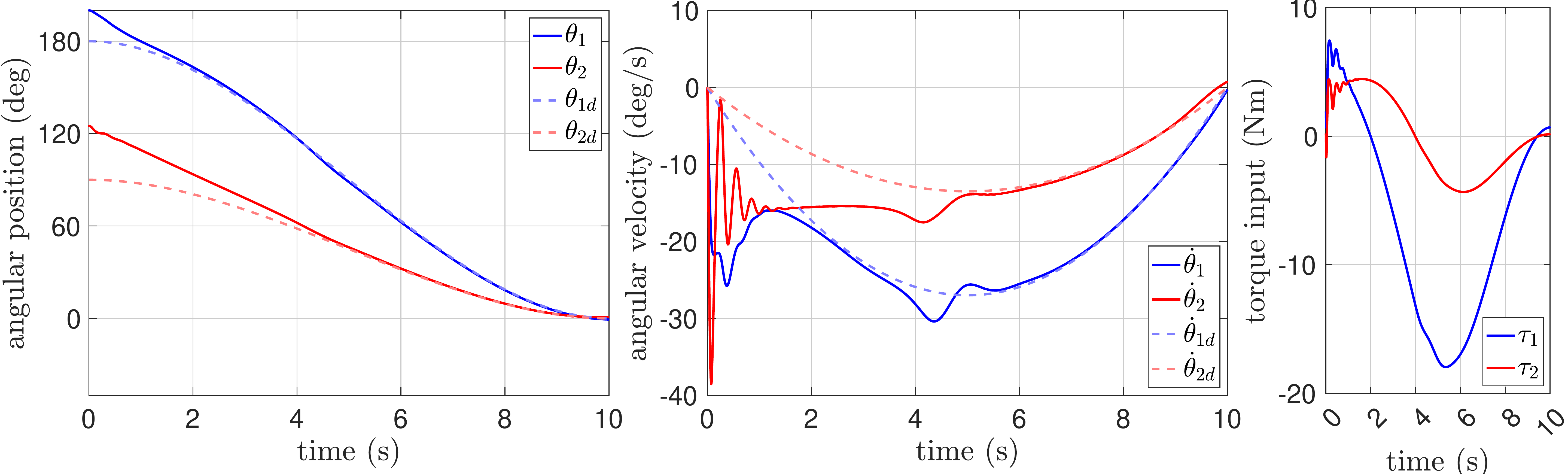}
    \caption{Motion trajectories and control inputs of the robot employing adaptive inverse dynamics control and the adaptation law. Despite considerable model uncertainties, the robot converges to the reference trajectory within approximately six seconds.}
    \label{fig:assignment_6b}
\end{figure}

The effectiveness of the adaptive control law is demonstrated through the plots for the performance of the robot, shown in Figures~\ref{fig:assignment_6a} and \ref{fig:assignment_6b}. In contrast to the nominal controller without an adaptive law, which fails to converge in the presence of significant model uncertainties, the adaptive control law enables the robot to converge to the reference trajectory and reach the desired upward configuration. By working on this assignment, students gain hands-on experience in developing an adaptive control design tailored to address model uncertainties and preserve tracking performance in robotic systems.


\FloatBarrier 

Video demos showcasing the robot's performance in the assignments presented in this section are available in the accompanying repository for this paper~\cite{sfarzan-git}, accessible at the following link: \\
\url{https://github.com/sfarzan/pbl_robot_control}

\section{Student Learning Outcomes}

In this section, we present the findings of the evaluations conducted to assess the effectiveness of the proposed project-based learning framework in the course ``Robot Control'', which is part of the Robotics Engineering Department curriculum at Worcester Polytechnic Institute. The course has been offered twice (in Spring 2022 and Fall 2022 semesters) with the current structure, and the evaluations are based on the course evaluation surveys filled out by the students at the end of the course, as well as feedback obtained from the students contacted after completing the course. A total of 50 students participated in the course evaluation survey for the Spring 2022 semester, while 37 students participated in the survey for the Fall 2022 semester.

The evaluation focused on assessing the success of the proposed project-based learning approach in enhancing students' understanding and application of control theory concepts, their programming skills, and their ability to apply the control algorithms and engineering skills in a practical setting. Additionally, the evaluation aimed to investigate the impact of the project-based learning approach on student satisfaction and engagement, retention and transfer of learning, and achievement of the learning objectives defined for the course.

The following subsections will provide detailed results of the course evaluations collected from the students, including both quantitative and qualitative results.
Throughout this section, we will highlight the positive feedback as well as suggestions for improvement provided by the students.

\subsection{End-of-course student survey}\label{subsec:survey}
By the conclusion of the course, the effectiveness of the project-based assignments was assessed using an anonymous end-of-course survey. Eight specific questions were asked to evaluate the success of the assignments on various aspects of the course. The survey used a 5-point Likert scale~\cite{joshi2015likert}, with options ranging from \textit{strongly agree} to \textit{strongly disagree}, as listed below:
\begin{center}
    \textit{Strongly agree (5) \quad Agree (4) \quad Neutral (3) \quad Disagree (2) \quad Strongly disagree (1)}
\end{center}

The questions aimed to evaluate the impact of the project-based assignments on enhancing the understanding of the course content by bridging the gap between theory and practice, improving programming skills, student satisfaction and engagement, retention and transfer of learning, and achieving the learning objectives of the course. The results of the survey were analyzed to understand the overall effectiveness of the project-based assignments in achieving the desired performance metrics for the course. The survey questions designed for evaluating the success of the assignments are listed below. 

\begin{itemize}
    \item[1.] The project-based assignments helped me connect the theory to applied practice.
    \item[2.] The project-based assignments were helpful in reinforcing the course materials covered in the lectures and enhance my understanding of the course content.
    \item[3.] The project-based assignments helped me improve my programming skills in MATLAB, Python, and Robot Operating System (ROS).
    \item[4.] The project-based assignments helped with keeping me motivated and engaged in the course content throughout the semester.
    \item[5.] The project-based assignments provided me with practical opportunities to apply my knowledge and skills, and were as valuable as the hands-on lab assignments typically found in other courses with labs.
    \item[6.] The project-based assignments helped with achieving the course learning objectives listed on the syllabus and discussed in the first lecture of the course.
    \item[7.] The knowledge and skills gained through the project-based assignments helped me to retain the course materials more effectively.
    \item[8.] The project-based assignments have prepared me for future careers in robotics and control engineering.
\end{itemize}

In addition to the end-of-course anonymous survey, the students who took the course in Spring 2022 were contacted in Fall 2022 to provide further feedback. They were asked to participate in a one-question anonymous survey to report whether they were able to apply the knowledge and skills learned in the project-based assignments to other courses or job interviews. The question asked was:\looseness=-1
\begin{itemize}
    \item[9.] I have been able to apply the knowledge and skills learned in the project-based assignments of the course to other courses and/or job interviews.
\end{itemize}

The survey results for both semesters during which the project-based assignments were implemented are listed in Table~\ref{tab:survery_results}.

\begin{table}[H]
    \centering
    \small
    \begin{tabular}{c c c c c}
    \toprule 
    \rowcolor{gray!25}
    Q\# & Performance Metric & Spring 2022 & Fall 2022 & Overall Score \\ 
    \midrule
    Q1 & Connecting theory and practice & 4.8 & 4.9 & 4.85 \\
    \rowcolor{gray!10}
    Q2 & Reinforcing the course content & 4.8 & 4.9 & 4.85 \\
    Q3 & Improvement in programming skills & 4.6 & 4.7 & 4.65 \\
    \rowcolor{gray!10}
    Q4 & Engagement with the course material & 4.7 & 4.7 & 4.7 \\
    Q5 & Applying knowledge \& skills in a practical setting & 4.7 & 4.8 & 4.75 \\
    \rowcolor{gray!10}
    Q6 & Meeting the learning objectives of the course & 4.8 & 4.8 & 4.8 \\
    Q7 & Improvement in ability to retain course material & 4.6 & 4.8 & 4.7 \\
    \rowcolor{gray!10}
    Q8 & Preparedness for future careers in robotics and controls & 4.6 & 4.6 & 4.6 \\
    Q9 & Transferability of the knowledge and skills & 4.2 & -- & -- \\
    \bottomrule
    \end{tabular}
    \normalsize
    \caption{Results of the end-of-course anonymous survey from students, showing the average scores for each performance metric evaluated for project-based assignments. All scores are out of 5.}
    \label{tab:survery_results}
\end{table}

In the remainder of this section, we will discuss the survey results presented above, analyzing the findings and drawing conclusions from the data. 

\subsection{Bridging the gap between theory and practice}

One of the primary objectives of incorporating project-based assignments was to assist the students in comprehending the theoretical concepts learned in the lectures, and to facilitate the bridging of the gap between theory and practice. Reinforcement of the content covered in lectures is an important consequence of successfully connecting theory and practice. Additionally, as other courses with labs typically offer hands-on lab assignments, we aimed to offer our students similar practical experiences to apply the knowledge and skills learned in the class. 

The student surveys included questions related to the students' understanding of the control theory concepts covered in the course and their ability to apply them in a practical setting (see Questions 1, 2, and 5 listed above). The survey results show that the
proposed project-based assignments have been effective in achieving these objectives.
Specifically, the overall scores ranging from 4.75/5.0 to 4.9/5.0 suggest that the students found the project-based assignments helpful in bridging the gap between theory and practice and gaining a better understanding of the theoretical concepts taught in the lectures by providing them with a practical setting to apply control concepts in real-world scenarios. The positive feedback from Question 5 is particularly noteworthy, as the physics-engine based simulation used in the project-based assignments has provided students with an experience comparable to an actual hardware lab, which may not be feasible for many institutions to implement due to the need of access to expensive hardware and resources. Students' qualitative course evaluations were generally positive and constructive in this regard, as shown in the feedback included below.

The surveys included comments from students, which revealed that they highly valued the hands-on experience provided by the assignments. According to their feedback, this experience helped them learn more effectively and improved their understanding of the course material:

\textit{``The assignments relation to the content taught in the course gave a better understanding of the concepts. The way assignments were dependent on previous assignment helped see inter-relation between different controls concepts.''}

\subsection{Improvement in programming skills}
The project-based learning assignments required students to develop control algorithms in MATLAB and Python for implementation in the ROS framework, a widely used software platform in robotics. To assess the impact of the assignments on students' programming skills, we analyzed the results of Question 3 in student surveys.
The survey results indicated that the project-based assignments had a positive impact on the students' programming skills in MATLAB and Python for ROS, with an overall score of 4.65/5.0.

This is a significant finding as programming skills are crucial for success in the field of robotics. The improvement in programming skills not only prepares the students for future courses in the robotics curriculum that require programming proficiency, but also enhances their preparedness for job interviews and future careers in the field. The ability to program robots is a highly valued skill in the industry, and the success of these project-based assignments in improving programming skills is a promising outcome for the students' future prospects in the field.

Furthermore, student survey results indicated that the assignments had helped them to learn and apply programming concepts more effectively, which they attributed to the hands-on experience provided by the assignments:

\textit{``The course taught me how to program robot manipulators and mobile robots using the Robot Operating System ROS. The assignments helped me improve my ability with MATLAB and Python programming languages.''} \\
\textit{``The way the ROS assignments were formulated, lead to a very good understanding of using the software that we can use in our future courses as well.''}

\subsection{Student satisfaction and engagement}
To measure the level of student satisfaction and engagement, the end-of-course survey consisted of several questions on students' overall satisfaction with the course, their level of engagement with the course material, and the extent to which they felt the course had prepared them for their future careers.

The survey results, especially the high overall score of 4.7/5.0 for Question 4, indicated a high level of satisfaction and engagement among the students with the project-based assignments. The vast majority of students reported in their qualitative comments that they found the course material, specifically the project-based assignments engaging and interesting, and that they felt they had learned a lot from the course. The assignments also gave them an opportunity to apply the concepts they had learned in lectures to real-world problems through modern robotics tools and technologies such as ROS and Gazebo, which they found engaging and motivating:

\textit{``This might have been my favorite course I have taken, and I have taken all of the undergraduate courses here as well. I really enjoyed how math-dense this course was, and also how it taught us how to apply theory to working controllers. I definitely felt like a masters degree student while taking this course; it was dense and taught a lot of really cool stuff.''}

\subsection{Achievement of learning objectives}
The proposed project-based learning approach has been successful in helping students achieve the learning objectives defined for the course, as previously listed in Section~\ref{sec:intro}. The project-based assignments have been designed to align with those learning objectives, and students have been able to use their theoretical knowledge and programming skills to develop control algorithms and apply them to simulated robots in real-world environments. The high overall score of 4.8/5.0 for Question 6 in the survey further implies the effectiveness of the project-based assignments in achieving the learning objectives in students' perspective.

Throughout the assignments, the students have been able to derive mathematical dynamic models for robotic systems using Lagrange's equations of motion, and analyze linear and nonlinear dynamical systems in terms of stability and convergence using mathematical principles. They have also been able to design model-based robot control algorithms using classical and modern control theory, and formulate advanced feedback control laws, including optimal, robust and adaptive algorithms, for controlling robots with unknown/uncertain parameters. All these achievements are closely aligned with the learning objectives set forth for the course.

Moreover, in addition to demonstrating a solid understanding of the mathematical concepts and applying them to solve practical problems in robotics, the students have been able to implement control algorithms on physics engine robot simulators to connect theory and practice, and construct, program, and evaluate the performance of robotic systems to perform specified control tasks. The project-based assignments have helped students bridge the gap between theoretical concepts and practical applications, which has been reflected in their grades and survey results.\looseness=-1

\subsection{Retention and transfer of learning}

To evaluate the retention and transfer of learning, we asked students to provide feedback on how the project-based assignments helped them retain course materials and prepare for their future careers (as indicated by Questions 7 and 8 listed in Section~\ref{subsec:survey}). In addition, to better understand the long-term impact of the assignments, a follow-up survey was conducted in Fall 2022 with students who had taken the course during the Spring 2022 semester (Question 9).

The survey results, specifically the overall scores of 4.7/5.0, 4.6/5.0 and 4.2/5.0 for Questions 7, 8, and 9 respectively, suggest that the project-based assignments not only helped students to better understand the course materials but also helped them to develop transferable skills that were applicable beyond the scope of the course. 
Moreover, according to the comments provided, students have indicated that the skills gained through the ROS-based project-based assignments have been transferable to other courses in the curriculum and capstone projects. 
Additionally, students reported that they have been able to use the skills in their interviews with companies, and they have been successful in securing job offers due to those skills.
This demonstrates the practical relevance of the proposed project-based assignments and their value in preparing students for their future careers in the field of robotics and control engineering.

\section{Conclusions and Future Directions}

In conclusion, the proposed project-based learning framework offers a valuable opportunity for students to gain practical experience in control theory for robotics using the Robot Operating System (ROS) and Gazebo simulation framework. By covering a comprehensive range of linear and nonlinear control concepts in the assignments, students are able to bridge the gap between theory and practice, and develop the necessary skills to tackle real-world challenges in robotics control. Furthermore, through the use of MATLAB, Python, and ROS, students can seamlessly apply their knowledge to practical assignments with the RRBot robot in Gazebo. This approach also aids students in enhancing their programming skills and ROS expertise, qualities that are in high demand within the robotics sector, leading to improved career prospects for the students. Our approach has been well received by students, as evidenced by positive course evaluations and surveys.\looseness=-1

Although our PBL framework has been successful in teaching a variety of control design techniques, there were certain course topics, including sliding mode control, force and impedance control, model predictive control, and control Lyapunov functions (CLFs), that could not be incorporated into the assignments due to time constraints.
To address this limitation, one potential improvement for future iterations of the course would be to split it into two distinct courses. By doing so, it would be possible to maintain a more reasonable pace while offering supplementary assignments that support the other topics discussed in the course.
Furthermore, students who lack prior experience with MATLAB and Python programming may encounter difficulties with the assignments. Dividing the course into two separate courses would enable the inclusion of introductory lectures on MATLAB, Python, and ROS in the first course, which would better equip students to handle the project-based assignments.
Overall, the proposed PBL framework offers an innovative and practical approach to teaching robot control theory, and there is scope for future improvements to further enhance the learning experience for students.

Moving forward, we believe that the proposed project-based learning framework can serve as a model for other courses looking to provide a more engaging and practical learning experience for students in the fields of control theory and robotics. 
For example, in addition to our PBL platform for postgraduate students in robot control theory, other robot models and packages available in ROS/Gazebo can be utilized for undergraduate level courses in controls. For instance, a self-balancing robot or a cart-pole system in Gazebo can be employed to teach transfer function based control designs, where students can derive and validate equations of motion, practice transfer functions, analyze transient and frequency responses, evaluate stability and performance of closed-loop systems, and ultimately stabilize the robot using a PID control design.

The proposed framework offers an effective project-based learning for undergraduate and graduate students to develop practical skills and gain hands-on experience in controls, which can be applied to a variety of fields in robotics and engineering.
By leveraging ROS and Gazebo's simulation environment, students can conduct real-world experiments and evaluate control designs without the need for expensive physical setups, thereby reducing costs and increasing accessibility to practical learning.
This approach can help bridge the digital divide, empowering students from remote or underprivileged areas to access cutting-edge technology and resources.
It also promotes equity in education, as it enables students from diverse backgrounds and financial circumstances to gain hands-on experience in robotics, regardless of their socio-economic status.

\vspace{3\baselineskip}\vspace{-\parskip}
\small
\bibliographystyle{IEEEtran}
\bibliography{ASEEpaper}

\end{document}